\pdfoutput=1

\documentclass{valence} %
\usepackage{soul} %
\usepackage{float}
\usepackage{amsmath}
\usepackage{algorithm}
\usepackage{algpseudocode}
\usepackage{amsfonts}
\usepackage{subcaption}
\usepackage{longtable}

\usepackage[export]{adjustbox}
\usepackage{comment}
\usepackage{mathtools}
\usepackage{authblk}
\usepackage{fancyhdr}
\usepackage[utf8]{inputenc}

\usepackage{amsmath,amsfonts,bm}

\def\eqref#1{equation~\ref{#1}}

\def\1{\bm{1}}

\DeclareMathAlphabet{\mathsfit}{\encodingdefault}{\sfdefault}{m}{sl}
\SetMathAlphabet{\mathsfit}{bold}{\encodingdefault}{\sfdefault}{bx}{n}

\newcommand{\rawmodelname}{TxPert}
\newcommand{\modelname}{\textit{\rawmodelname}}

\newcommand{\exphormermodelname}{Exphormer}
\newcommand{\exphormermodel}{\emph{\exphormermodelname}}

\newcommand{\exphormerMGmodelname}{Exphormer-MG}
\newcommand{\exphormerMGmodel}{\emph{\exphormerMGmodelname}}

\newcommand{\hybridmodelname}{GAT-Hybrid}
\newcommand{\hybridmodel}{\emph{\hybridmodelname}}

\newcommand{\multilayermodelname}{GAT-MultiLayer}
\newcommand{\multilayermodel}{\emph{\multilayermodelname}}

\title{\rawmodelname: Leveraging Biochemical Relationships for Out-of-Distribution Transcriptomic Perturbation Prediction}

\author[1,2,\dagger,\star]{Frederik Wenkel}
\author[1,2,\dagger]{Wilson Tu}
\author[1,2]{Cassandra Masschelein}
\author[1,2]{Hamed Shirzad}
\author[1,2]{Cian Eastwood}
\author[1,2]{Shawn T. Whitfield}
\author[1,2]{Ihab Bendidi}
\author[1,2]{Craig Russell}
\author[1,2]{Liam Hodgson}
\author[1,2]{Yassir El Mesbahi}
\author[3]{Jiarui Ding}
\author[2]{Marta M. Fay}
\author[1,2]{Berton Earnshaw}
\author[1,2]{Emmanuel Noutahi}
\author[1,2,\star]{Alisandra K. Denton}

\affiliation[1]{Valence Labs, Montr\'eal, QC, Canada}
\affiliation[2]{Recursion, Salt Lake City, UT, USA}
\affiliation[3]{Computer Science, University of British Columbia, Vancouver, BC, Canada}
\contribution[\dagger]{These authors contributed equally}
\contribution[\star]{\small Correspondence: frederik@valencelabs.com, ali@valencelabs.com}

\abstract{
Accurately predicting cellular responses to genetic perturbations is essential for understanding disease mechanisms and designing effective therapies. Yet exhaustively exploring the space of possible perturbations (e.g., multi-gene perturbations or across tissues and cell types) is prohibitively expensive, motivating methods that can generalize to unseen conditions. In this work, we explore how knowledge graphs of gene-gene relationships can improve out-of-distribution (OOD) prediction across three challenging settings: unseen single perturbations; unseen double perturbations; and unseen cell lines. In particular, we present: (i) \modelname, a new state-of-the-art method that leverages multiple biological knowledge networks to predict transcriptional responses under OOD scenarios; (ii) an in-depth analysis demonstrating the impact of graphs, model architecture, and data on performance; and (iii) an expanded benchmarking framework that strengthens evaluation standards for perturbation modeling.  
}

\begin{document}

\maketitle

\section{Introduction}\label{sec:intro}
Living systems are highly complex, specialized, and varied. To account for this complexity, candidate therapeutics are tested across a range of assays in diverse cellular contexts before advancing to models such as organoids, xenografts, animals, and ultimately human clinical trials. Yet, the vast majority of these candidate therapeutics are ultimately unsuccessful, often failing late in development after considerable costs have been incurred. At its core, a therapeutic is a perturbation intended to shift a cell from one state to another one, typically from a disease state to a healthier one. Finding the perturbation that will produce the desired effect is therefore central to drug discovery. However, due to cost and scale, experimental screens can only cover a small fraction of possible perturbations and contexts, making prioritization essential. As a result, there is a growing need for computational models that can simulate the effects of perturbation \emph{in silico} across biological contexts. Such models would reduce the need for exhaustive screening, enable principled extrapolation across cell types and conditions, and accelerate the design of effective therapeutic interventions.

 Two complementary strategies have emerged to predict perturbation effects in a generalizable, out-of-distribution (OOD) setting. The first strategy exploits the inherent compositional nature of cellular responses by training deep generative models to learn latent representations that can be perturbed, enabling \textit{in silico} predictions across contexts ~\citep{lotfollahi2019scgen, cpa}. 
 The second leverages prior biological knowledge, either by incorporating curated gene–gene relationship networks or embedded functional annotations, to impart strong inductive biases that improve generalization \citep{sclambda, gears}. Most existing models for perturbation effect prediction are trained and evaluated on a per-dataset basis, limiting their ability to generalize across biological contexts or scale in real-world application. This restricts their utility in drug discovery settings, where performance across unseen perturbations, cell contexts, and experimental conditions is essential. 

While machine learning models have made major strides in some fields of biology like protein structure prediction \citep{jumper2021highly}, deep models for transcriptomics-focused perturbation biology have often underperformed, sometimes trailing simple or even untrained baselines~\citep{ihab_benchmark, wong2025simple, ahlmann2024deep, perturbench}.
Notably, \citet{kernfeld2023systematic} showed that the training set mean or median of all perturbation responses is a better predictor of an individual perturbation response than the outputs from prominent models like GEARS \citep{gears}. A historical absence of 
basic baselines in primary modeling papers has hindered both fair model comparison and iterative improvement. In short, the potential of deep learning models in this biologically important domain remains an open research question. 

In this work we introduce \modelname~as a unifying model for transcriptomics perturbation effect prediction that (1) can be trained broadly across datasets, (2) supports three major out-of-distribution tasks in a single framework, and (3) effectively leverages prior knowledge and biological context without requiring dataset-specific optimization.
In particular, we show that \modelname~achieves state-of-the-art performance in predicting unseen single perturbation effects within cell types and for OOD transfer to unseen cell lines, while showing highly competitive performance in predicting the effects of double-gene perturbations.

Beyond \modelname, we present a modular and extendable training and evaluation framework for transcriptomic perturbation prediction that advances best practices with rigorous benchmarking. Specifically, we introduce batch-appropriate control matching; stratified performance metrics across more- and less-characterized areas of biology; and recently introduced evaluation tasks such as retrieval \citep{perturbench,szalata2024benchmark}. Finally, we contextualize the model's performance with comparisons to multiple baseline models and published methods, as well as estimates of experimental reproducibility.

\section{Results}

\modelname~is a deep learning framework designed to predict the transcriptional response to single or combinatorial genetic perturbations, both within and across cell types. Before addressing modeling considerations, we first re-examined the appropriateness of common data handling practices and evaluation metrics to ensure they align with the statistical structure and biological properties of perturbation data.

\subsection{Revisiting Metric Design for Biologically-grounded Perturbation Effect Modeling}
Modeling of transcriptomic cell profiles after a perturbation is a fundamentally nascent field with many open questions and a notable absence of clear consensus on best practice for data handling and evaluation. 
Given this uncertainty, we first conducted a theory-informed and data-driven investigation to 
better understand the data being modeled so that we could maximize the biological
impact of our results. 

\begin{figure}[tb]
  \centering
  \includegraphics[width=\textwidth]{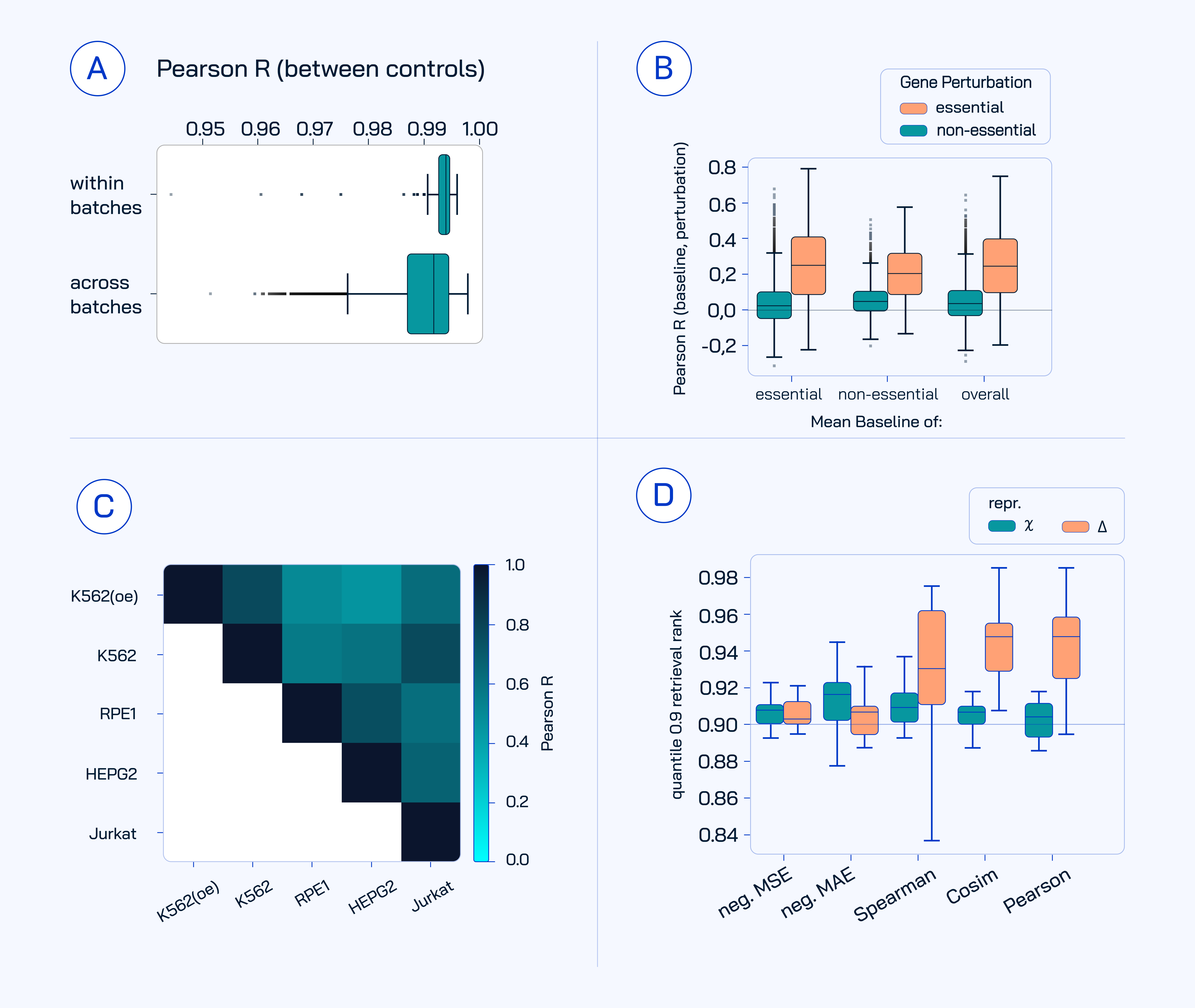}
  \caption{
A) Pearson correlation of aggregated control gene expression profiles within and across experimental batches. 
B) Correlation between single perturbations and the mean baseline, i.e., the mean delta calculated over aggregates of essential (as defined by \citet{replogle}), non-essential, or all genes ($\#\text{samples} \, \in\{2058, 7815, 9866\}$, respectively). 
C) Correlation between the mean baseline aggregated within or between studies and cell types. All data are CRISPRi unless marked with (oe) for overexpression data from \citet{norman2019exploring}. 
D) Normalized retrieval between true perturbant replicates in different biological contexts. Retrieval is calculated based on the indicated expression representations and metrics ($\#\text{samples} \, = 18$). The plotted value is the 0.9 quantile (across all unique perturbants), where expected random performance is 0.9, indicated by the dashed line.
}

  \label{fig:dataanalysis}
\end{figure}

\textit{\textbf{Batch-matched controls are warranted given batch effects and confounding in the datasets}}

Experimental batch effects are a well-known challenge in  biological data~\citep{replogle, celik2024building} where absolute values vary drastically due to the sensitivity and variability of biological systems and their interaction with the environment. To account for this variability, primary experimental studies include carefully designed controls in every batch. However, many deep learning models \citep{gears} rely on a global control mean to compute perturbation effects (`delta' expression metrics, see Section~\ref{sec:method}). 
In a scenario where batch effects are present and any confounding between batch and applied perturbation occurs, failure to account for the batch effects can result in mistaking background batch-wise variance for perturbation effect and overestimating model performance, reduce true performance by adding avoidable noise, or both. We confirmed the substantial batch effects by analyzing the genome-wide Perturb-seq dataset from~\citet{replogle}, where we quantified control–control correlations both within and across batches. Across-batch correlations were significantly lower (U-test, $p=1.4 \times10^{-19}$) despite involving more aggregated cells (Fig.~\ref{fig:dataanalysis}A). Subsequently, we checked for confounding effects between batch and perturbation. As a library of perturbants is applied to pooled cells in PerturbSeq, and all of transfection, survival, and sampling are stochastic processes, confounding cannot be strictly controlled but it can be measured. We found significant association between batch and perturbation ID in every dataset used here ($\tilde{\chi}^2$, $1.1 \times 10^{-4}$ to $1.0 \times 10^{-98}$ across datasets).
These observations confirm significant batch effects and confounding between perturbation identity and batch assignment. Consequently, to maintain biological validity and reduce noise, we adopted batch-matched controls for all subsequent model training and evaluation.

\textit{\textbf{A deep dive into the data justifies focusing on retrieval metrics and Pearson $\Delta$}}

Prior studies have shown that mean-baseline models (averaging perturbation effects across many perturbations) achieve surprisingly strong predictive performance~\citep{kernfeld2023systematic, wong2025simple}.
We found that this is especially prominent in perturbations of ``essential'' genes (Fig.~\ref{fig:dataanalysis}B) and related to general cellular responses. Consistent with the idea that a perturbed cell undergoes general stress and shifts from growth and uptake to quiescence and recycling, for the mean perturbation effect we observed an enrichment of  the functional terms \textit{vacuole}, \textit{autophagosome membrane}, and \textit{lysosome} among upregulated genes, and functions related to cell division and various metabolic processes among downregulated genes (Tab.~\ref{tab:go_mb_up},\ref{tab:go_mb_down}).
Importantly, we found that these baseline effects were consistent across diverse datasets and perturbation modalities, including
the \citet{norman2019exploring} data, which used a variant of CRISPR, 
CRISPRa, to upregulate target genes, in contrast to the other studies using 
CRISPRi to downregulate targets (Fig.~\ref{fig:dataanalysis}C). In short, 
the strong predictive power of the mean baseline is a feature of the data. It reflects general biological responses to perturbation-induced stress or a reduction
in fitness, health or growth due to perturbation of an important cellular component, rather than perturbation-specific effects.

Given these findings, evaluating models purely based on mean perturbation responses is insufficient for assessing their capacity to distinguish between biologically distinct perturbations. 
We therefore adopted a complementary evaluation approach using retrieval metrics, which explicitly measure how effectively a model distinguishes replicate perturbations from other perturbations~\citep{perturbench,szalata2024benchmark}. Retrieval performance depends on both (1) the representation of the perturbation effects and (2) the similarity metric chosen, and the choices
of these factors vary considerably in the field. Through systematic benchmarking, we found that cosine similarity and Pearson correlation applied directly to delta profiles yielded the highest retrieval scores (Fig.~\ref{fig:dataanalysis}D). On the other hand, common practices such as selecting only the top differentially expressed genes substantially reduced retrieval performance compared to unfiltered Pearson $\Delta$ (Fig.~\ref{fig:no_top_DE}).  Consequently, we selected Pearson $\Delta$ as our primary evaluation metric, complemented by retrieval scores during final testing to holistically assess perturbation-specific signal retention.

\begin{figure}[tb]
  \centering
  \includegraphics[width=\textwidth]{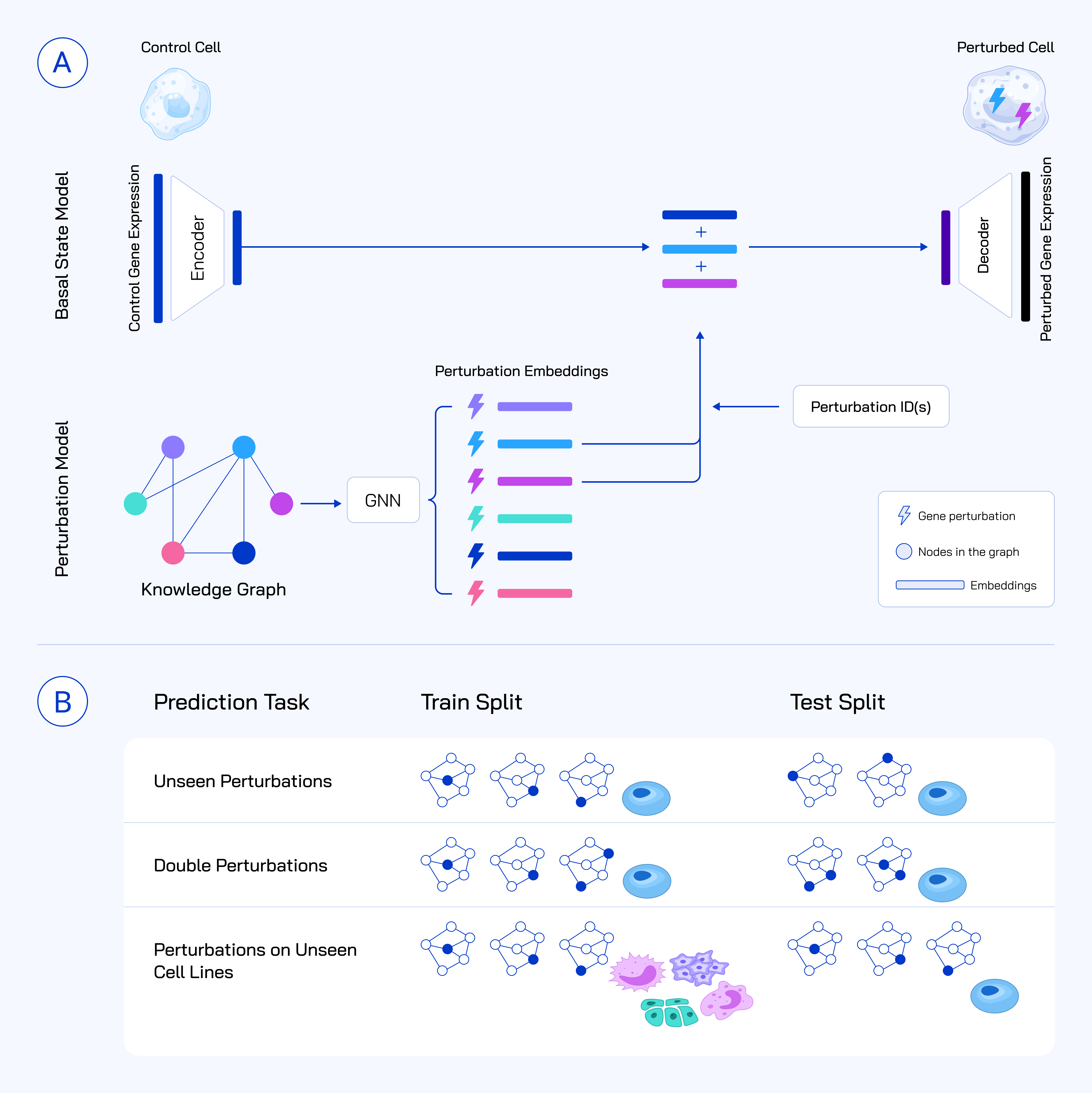}
  \caption{A) The \modelname~architecture predicts post-perturbation gene expression by combining two modules: (1) a basal state encoder that maps batch-matched control profiles into a latent embedding, and (2) a Graph Neural Network (GNN) that learns perturbation embeddings from a gene-gene interaction graph. Perturbation embeddings are applied to the basal embedding, and the resulting latent representation is decoded to produce the predicted gene expression profile.
 B) OOD Perturbation effect prediction tasks for: (i) unseen single perturbations within the training cell line, (ii) novel double perturbations, where constituent singles may have been seen during training, within the training cell line, and (iii) perturbations within new cell lines not seen during training.}
  \label{fig:architecture}
\end{figure}

\subsection{\rawmodelname: A framework for predicting perturbation effects on transcriptomics across multiple OOD tasks}

We introduce \modelname, a deep learning framework designed for robust prediction of transcriptional responses to previously unobserved genetic perturbations, including single-gene perturbations, combinations of perturbations, and perturbations across new cell types.  Inspired by previous efforts~\citep{cpa, sclambda}, \modelname~relies on the \emph{latent transfer} paradigm to achieve strong generalization performance. Specifically, it integrates two complementary modules:   a \emph{basal state encoder} that learns an embedding of the cell prior to perturbation (i.e., cell type, protocol, batch, etc.), and a \emph{perturbation encoder} that learns a representation of perturbation(s) by leveraging informative embeddings from gene interaction networks (Fig.~\ref{fig:architecture}). These embeddings are combined (latent transfer), and decoded to predict the resulting log-transformed gene expression profile of perturbed cells.

\begin{figure}[tb]
  \centering
    \includegraphics[width=\textwidth]{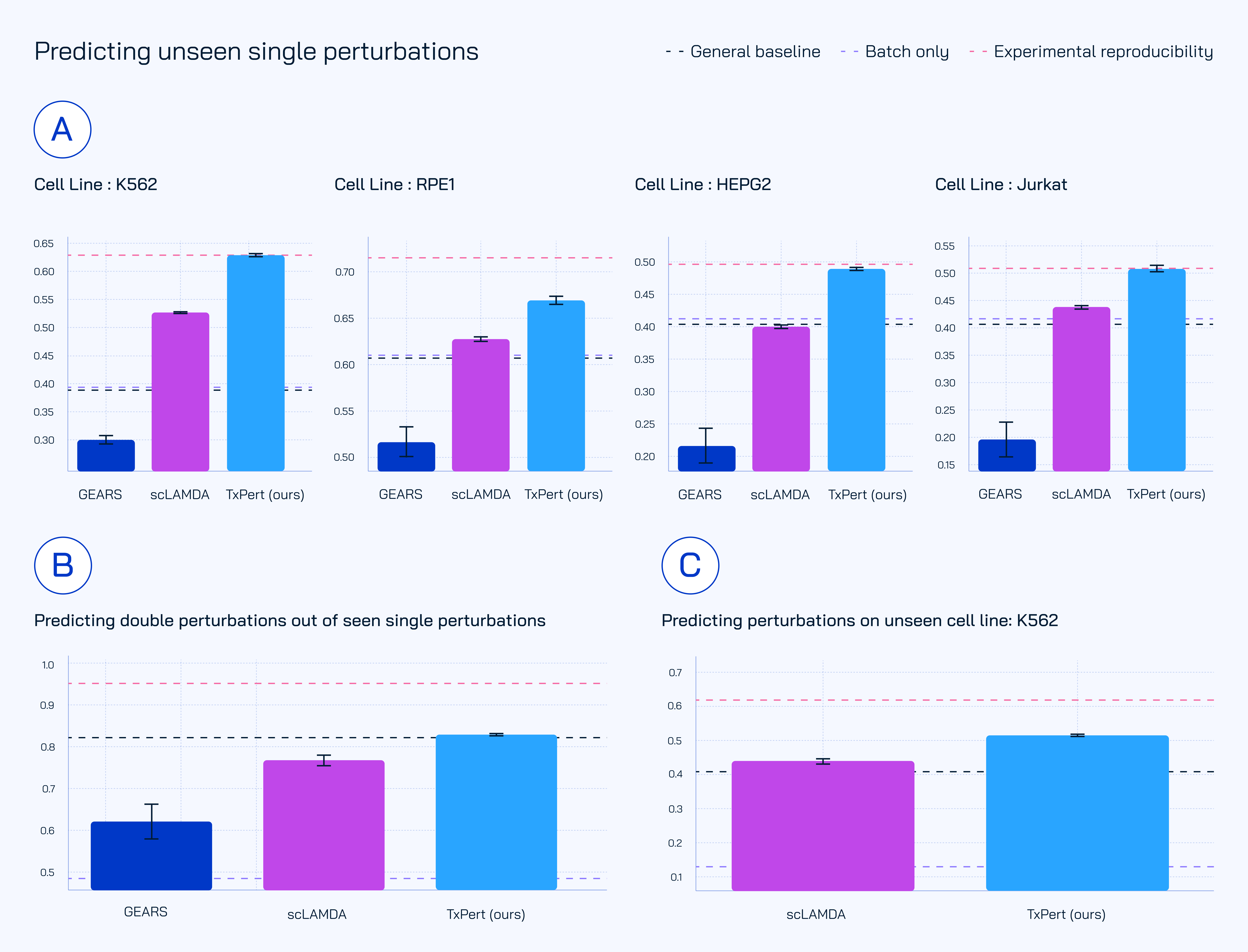}

  \caption{A) Performance of \modelname~compared to GEARS and scLAMBDA on predicting unseen single perturbations within a known cell type. Horizontal bars indicate general baseline, a batch-informed model (capturing potential confounding), and experimental reproducibility (see Section~\ref{sec:method}). B) Comparison of \modelname, GEARS and scLAMBDA in predicting double perturbation effects from known singles. C) Comparison of \modelname~and scLAMBDA performance on predicting single perturbations in unseen cell lines.}
  \label{fig:sota_unseen}
\end{figure}

Through extensive exploration of candidate methods for each module across multiple datasets, we identified architectural choices that effectively leverage existing biological knowledge for generalizable perturbation prediction. For the basal state model, we explored encoding gene expression counts using (1) a multilayer perceptron and (2) pretrained embeddings from established foundation models. For the perturbation model, we investigated (1) major graph neural network architectures (graph attention networks, graph transformers), as well as (2) hybrid and multilayer GNN variants that combine different gene interaction graphs~(see Section~\ref{sec:method}).

While all proposed GNN variants demonstrate strong performance across the explored OOD tasks, we tuned each architecture individually and report the best model variant per task in the main text. We provide further information on model choice after discussing architecture-specific details (Section~\ref{sec:ablations}). For the prediction of unseen single perturbations, we ultimately selected the \exphormerMGmodel~\citep{shirzad2023exphormer,shirzad2024even} graph transformer architecture, which enables us to simultaneously integrate four complementary graph sources: STRINGdb~\citep{string}, GO~\citep{go}, PxMap and TxMap. PxMap and TxMap are proprietary Recursion relationship-datasets derived from large-scale phenomics (microscopy imaging; \cite{cellpaint}) and single-cell transcriptomics perturbation screens, respectively~\citep{celik2024building,ph1}. Similarly, for predicting double perturbations, the \multilayermodel~\citep{zangari2024link, yun2021neo} achieves the best results, leveraging the complementary information from GO, PxMap and TxMap. Regardless of the specific architectural choices, our proposed framework consistently outperforms existing methods such as GEARS, scLAMBDA, and a general baseline, which predicts perturbation effects based on the strongest combination of additive and mean as appropriate for the task (see Section~\ref{sec:eval-general_baseline}). This superior performance is observed across various challenging out-of-distribution prediction tasks, demonstrating the framework's robustness and versatility.

\subsection{\rawmodelname~substantially outperforms other models at predicting unseen perturbation effects across known cell types}

Even in the simplest case of predicting the transcriptional response to an unobserved perturbation in a known cell type, achieving generalization requires relying heavily on contextual information, including perturbant-specific and biomolecular interaction signals learned from perturbations observed during training.
In this setting, we focus on four well-studied cell lines, each with extensive perturbation data publicly available (more than 2000 perturbation types per cell line): myelogenous leukemia lymphoblast (K562), retinal pigment epithelial cells (RPE1), liver hepatocellular carcinoma (HEPG2) and human T lymphocytes (Jurkat) \citep{replogle,nadig}. We trained \modelname~and existing baseline methods on each cell line data separately, while leaving out specific perturbations for evaluation.
We compare our models results to two existing methods for leveraging biological data. The first, GEARS \citep{gears}, 
integrates prior biological knowledge by embedding a gene–gene relationship network derived from Gene Ontology (GO) terms with a GNN architecture, and embedding basal state with a co-expression derived graph and GNN. The second, scLAMBDA, 
uses inductive biases derived from external textual data. Specifically, it leverages genePT—embeddings generated by applying GPT-3.5 to the functional summaries available from the NCBI gene database~\citep{sclambda}. The model maps each gene into disentangled vectors that reflect both its molecular function and its biological context. 
As illustrated in Fig.~\ref{fig:sota_unseen}A, \modelname~uniformly outperforms scLAMBDA, GEARS, and the general baseline, with GEARS notably falling below the non-learned general baseline.  More interestingly, our model is competitive with experimental reproducibility\footnote{Experimental reproducibility represents the reproducibility achievable by splitting the test set replicates, grouped by batch, in half and comparing the halves. Although not an absolute upper bound, it serves as a rigorous benchmark comparable to human-level performance in vision tasks.} in two out of the four cell lines (K562, Jurkat)
and comes close in a third (HEPG2). Retrieval metrics revealed similar relative performance patterns between models with \modelname~again outperforming other methods. However, the remaining gap between experimental reproducibility and \modelname~is much larger when measured by retrieval. In addition, the General Baseline has shown the weakest performance out of all approaches (Fig.~\ref{fig:sota_unseen_retrieval}).

\subsection{\rawmodelname~outperforms existing models, as well as the additive baseline in predicting the effect of multi-gene perturbations}

Although genome-wide single perturbation datasets are becoming increasingly available, combinatorial perturbation experiments remain prohibitively costly even for well-resourced labs, heightening interest in computational predictions.
We compare our model's performance at predicting the effect of double perturbations using the \citet{norman2019exploring} dataset, specifically focusing on cases where both individual perturbations had previously been seen in isolation during training. 
At this task, \modelname~achieves a remarkably higher Pearson $\Delta$ %
than GEARS and scLAMBDA, and outperforms the Additive Baseline\footnote{The additive baseline is a specific case of the proposed General Baseline, namely where the expression profile of the unseen double $(i, j)$ is predicted as $\hat{\mathbf{y}}_{(i,j)} = \bar{\mathbf{x}} + \delta_i + \delta_j$ with an appropriate (mean) estimate of the control in the test set $\bar{\mathbf{x}}$, e.g., aligned w.r.t. cell line and batch effect of the target.}
as seen in Fig.~\ref{fig:sota_unseen}B.  %

\subsection{\rawmodelname~generalizes effectively to predict perturbation effects across unseen cell lines}
The third key task explored here is predicting the perturbation effect of a seen perturbation in a new biological context (here, a new cell line). Predicting  perturbation responses in a new biological context represents a critical test of a model’s generalization capabilities, as substantial perturbation data exist only for a small fraction of cellular contexts. We assessed this challenging task through four leave-one-out experiments, where we hold out all perturbation examples from the target cell type, but do train on all controls. Neither GEARS nor scLAMBDA originally included this task,  however, we adapted scLAMBDA's implementation to provide a relevant baseline.
While our training data is large, it has extremely limited cell lines diversity, 
and when approaching this 0-shot generalization task, it was unclear what level
of performance was to be expected. Nevertheless, we found that \modelname~exceeded the general baseline in all four held out cell lines  and consistently outperformed scLAMBDA (Fig.~\ref{fig:sota_unseen}C, Fig.~\ref{fig:xcell_sup}).

Collectively, these results serve as a proof-of-concept, demonstrating that \modelname~can successfully address diverse out-of-distribution perturbation prediction tasks. It excels particularly in predicting unseen perturbations, while setting a competitive baseline for improvement in cross-cell line predictions.

\subsection{\rawmodelname~learns meaningful information from biological knowledge graphs}

\begin{figure}[tb]
  \centering
  \includegraphics[width=\textwidth]{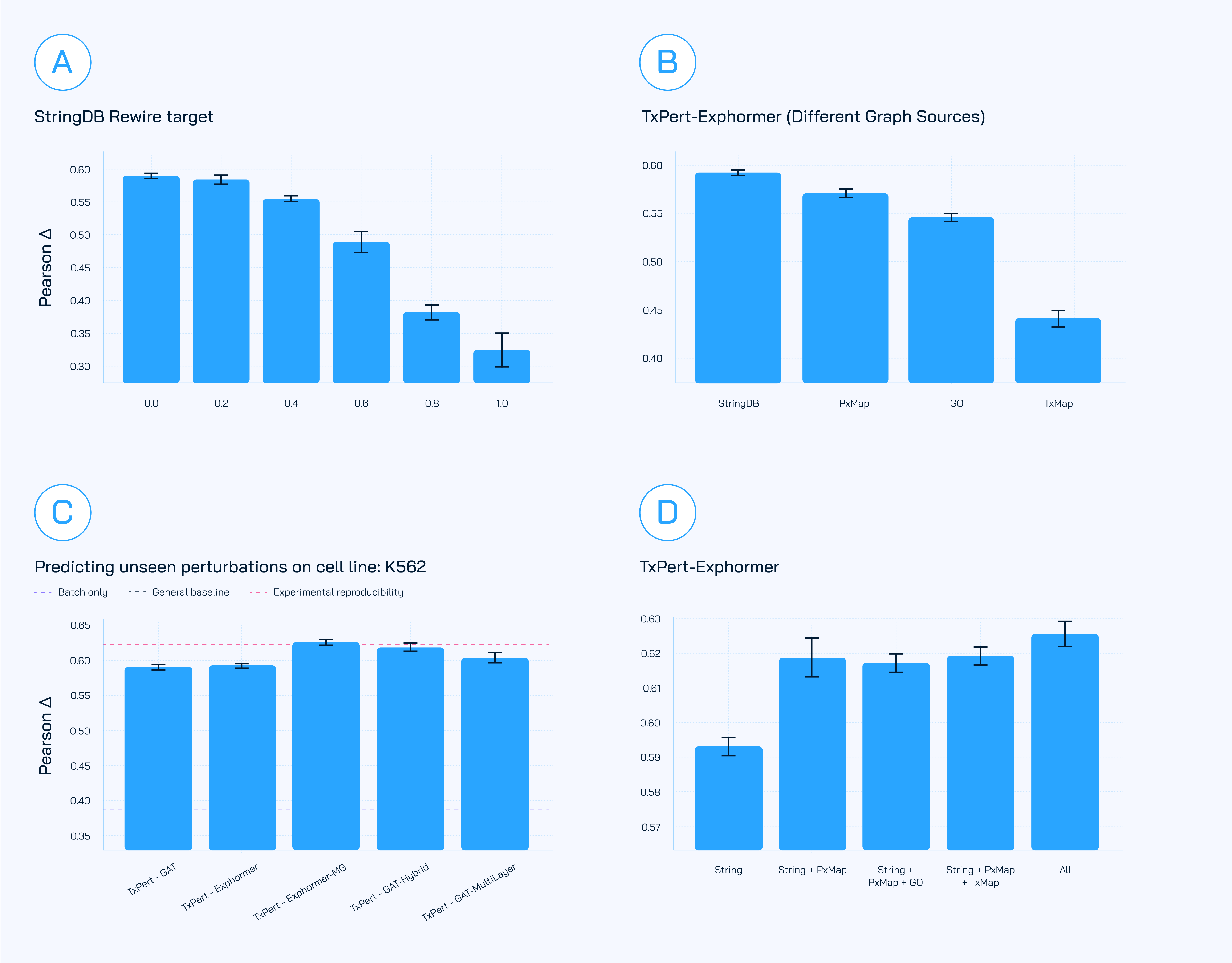}

  \caption{Ablation studies for unseen perturbation effect prediction on K562. A) Performance of \modelname~as edges of the STRINGdb graph are progressively rewired.
    B) Performance of \modelname~(Exphormer) using individual graphs.
    C) Comparison of various graph integration strategies and architectures. 
    D) Performance of \modelname~(Exphormer) as multiple knowledge graphs (STRINGdb, GO, PxMap, TxMap) are subsequently integrated into the Exphormer architecture (Exphormer-MG).
    Horizontal bars indicate general baseline performance, the performance of a learned model making predictions on the basis of batch information (in case of confounding between batch and perturbation), and an experimental
    reproducibility estimate.}
  \label{fig:graph_ablation_main}
\end{figure}

To understand how biological knowledge encoded in the gene interaction graphs contributes to predictive performance, we systematically assessed the utility of various graph priors.
Since no definitive, universally accepted biological interaction graph exists, we first empirically compared several graph sources. We evaluated two curated-database derived graphs (STRINGdb and GO) alongside two graphs derived from genome-wide perturbational screens (PxMap and TxMap). Among these alternatives, the \exphormermodel~configuration of \modelname~performed best with the STRINGdb graph (an accumulation of biological knowledge from database, literature and multiple raw sources) followed by the PxMap (derived entirely from high throughput perturbational screening in Primary Human Umbilical Vein Endothelial Cells; HUVEC)  (Fig.~\ref{fig:graph_ablation_main}B). 
Next, to determine whether the graph was fundamentally necessary, we evaluated the impact of progressively degrading its structure. Specifically, we randomly rewired the original STRINGdb graph (by changing the source, target, or both nodes of each edge) from 0\% (original graph) up to 100\% (completely randomized), retraining and assessing model performance at each step. As the proportion of randomized edges increased, we observed a consistent decline in the test set Pearson $\Delta$ for the K562 dataset (Fig.~\ref{fig:graph_ablation_main}A,\ref{fig:downsample_rewire}). Similarly, performance was also sensitive to randomly downsampling edges (Fig.~\ref{fig:downsample_rewire}). Surprisingly, models were relatively robust to downsampling (i.e., removing) up to 60\% of the edges. We hypothesize that the GNN architectures can compensate for this graph corruption by relying on multi-hop reasoning (as the model uses 4 GNN layers). To validate this, we perform another experiment, where we compare a 4-layer GAT architecture (selected via hyperparameter tuning) to a GAT only using a single layer (Fig.~\ref{fig:4vs1-layer-gnn}). Indeed, we observe a significantly stronger decline in performance as we downsample the graph for the one-layer architecture that cannot rely on multi-hop message passing. This observation is also in line with the properties of the STRINGdb graph that is used for the experiment having a diameter of four, i.e., every node being at most four hops from every other node in the graph. Together, these findings indicate that while predictions can be made without biologically informed priors, incorporating accurate biological graph information substantially enhances model performance.

\subsection{\rawmodelname~performance increases when combining multiple knowledge graphs}
We hypothesized that different biological graphs might provide complementary information and that their combination could yield improved prediction. To this end, we explored three different strategies for integrating multiple graphs: (1) \hybridmodel~, an extension of the GATv2 model designed to learn from several knowledge graphs simultaneously and subsequently combine their information; (2) \exphormerMGmodel~, an extension of the Graph Transformer architecture adapted for multi-graph learning via a union graph methodology; and (3) \multilayermodel~, a method that adapts GATv2 to operate on a unified supra-adjacency representation of multiple knowledge graphs, enabling message passing across both intra-layer and inter-layer connections simultaneously. Detailed descriptions of these models are provided in Appendix~\ref{apx:gnns}. For simplicity we focused on predicting unseen perturbations in K562 cells, and here the Hybrid Graph Transformer, (\exphormerMGmodel), achieved the highest performance (Fig.~\ref{fig:graph_ablation_main}C). 
Moreover, incremental integration of multiple graphs starting from STRINGdb consistently improved predictive performance, peaking when all four graphs (STRINGdb, GO, PxMap, TxMap) were combined (T-test, $p < 0.027$ when comparing best three-source graph vs four-source) (Fig.~\ref{fig:graph_ablation_main}D). 

\subsection{Detailed evaluation of model performance}
To understand our model's performance in more depth than just global metrics, we undertook a detailed analysis and scrutinized factors that could relate to model performance. 

\begin{figure}[tb]
  \centering
  \includegraphics[width=0.8\textwidth]{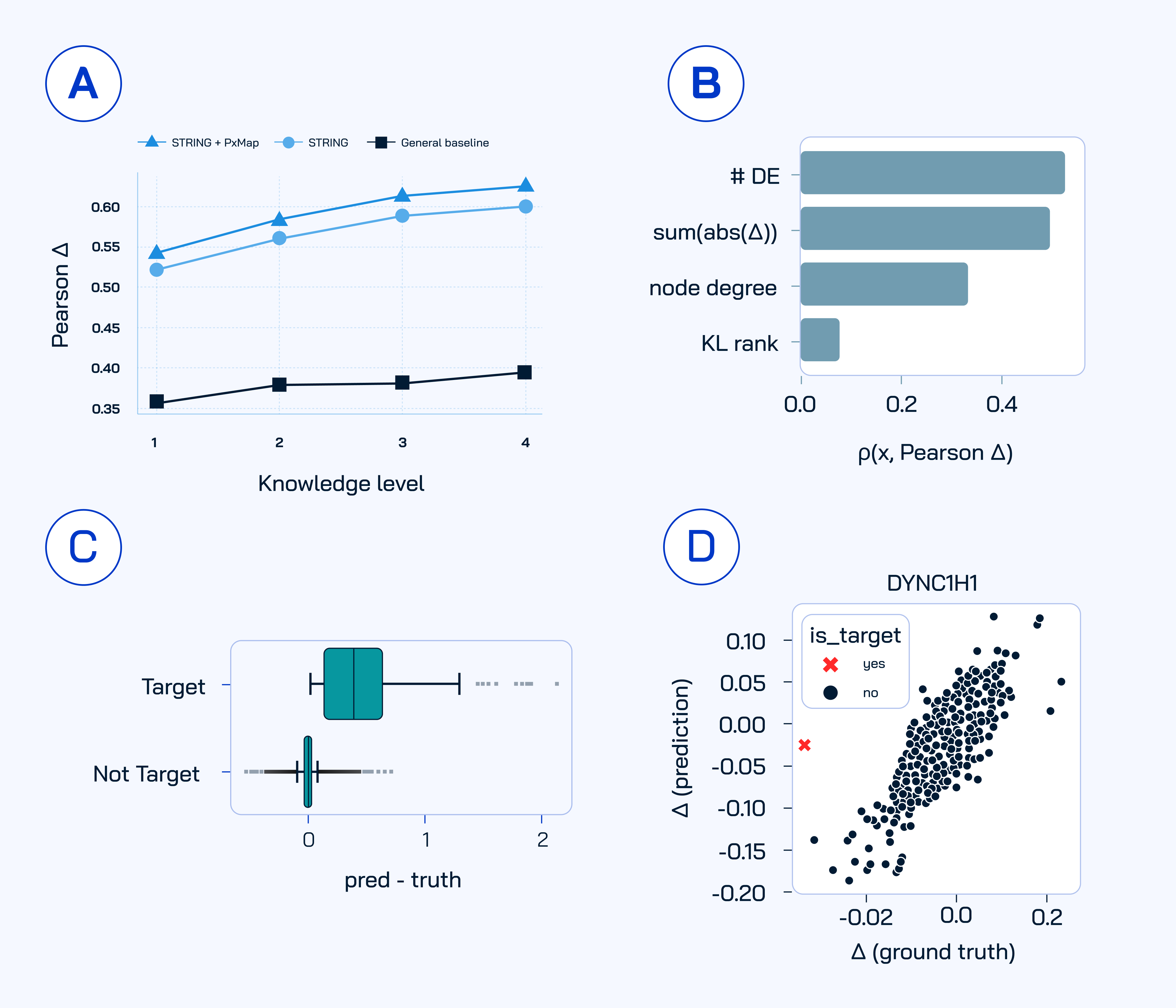}
  
  \caption{Investigation into strengths and weaknesses of our models. A) Breakdown of Pearson $\Delta$ by the knowledge level (Pharos rank) of the assayed genes. B) Spearman correlation between performance (Pearson $\Delta)$ and both data intrinsic factors (number of differentially expressed genes, sum of absolute $\Delta$s) and biological knowledge factors (degree of perturbed node in graph, Pharos knowledge level) metadata for unique perturbation which were hypothesized to be related to performance. C) Signed error in predicting the expression of perturbation targets, when these either are or are not targets. D) Example prediction vs ground truth for all genes when DYNC1H1 is perturbed, showing the target, DYNC1H1 in red. DYNC1H1 was chosen as an arbitrary but representative example demonstrating the common failure to predict the true down regulation of the perturbation target.}
  \label{fig:performance_deep_dive}
\end{figure}

First, as our model is leveraging prior information, which we know to some degree is both incomplete and biased, we investigated if there were any patterns in how the knowledge available about a target or predicted gene related to prediction accuracy. 
We binned expressed genes into four knowledge levels by their Pharos knowledge rank (\cite{sheils2021pharos} and see Section~\ref{sec:data}), and calculated Pearson $\Delta$ separately on these subsets. Unsurprisingly, we had higher prediction error for genes that have a lower rank and less knowledge about them available. However, so did the General Baseline, indicating that \emph{a portion} of this pattern is driven by gene-intrinsic factors such expression level and responsiveness to perturbation. We hypothesized that the perturbation-effect derived ``maps'' would show less bias, and encouragingly found that a hybrid graph where we added the PxMap to the STRINGdb graph showed consistent improvement across all knowledge levels (Fig.~\ref{fig:performance_deep_dive}A). 

Next, we examined relations between information about the perturbation target and prediction quality from a \hybridmodel~model (STRINGdb and TxMap). 
We tested hypotheses specific to our modeling strategy leveraging graphs and prior biological knowledge, 
and found that curated graph degree has a significant Spearman's correlation with Pearson $\Delta$ (Fig.~\ref{fig:performance_deep_dive}B, \ref{fig:extra_eval}C; $\rho = 0.33$, $p = 0.00062$); however we observed
no significant correlation between the Pharos knowledge
rank of a perturbation target and the resulting Pearson $\Delta$ (Fig.~\ref{fig:performance_deep_dive}B, \ref{fig:extra_eval}D; $\rho=0.074, p=0.22$); 
note this differs from the relation observed between knowledge rank and error by \emph{assayed gene}, above. In contrast, clear relationships
were found regarding the overall 
perturbation effect size as estimated by the count of differentially expressed genes, or the sum of the absolute deltas, respectively, 
had a greater correlation (Fig.~\ref{fig:performance_deep_dive}B, \ref{fig:extra_eval}A, $\rho=0.52$, $p=2.5 \times 10^{-20}$; Fig.~\ref{fig:performance_deep_dive}B, \ref{fig:extra_eval}B, $\rho=0.49$, $p=4.0\times10^{-18}$), potentially because a stronger perturbation effect has a higher signal to noise ratio.
Running functional enrichment on the perturbations with highest and lowest Pearson
delta scores, we found strong enrichments in what our model excelled at, such as protein translation and localization, whereas there were no significant enrichments ($p > 0.05$)
in the perturbants our model performed worst on (Tab.~\ref{tab:go_targ_best}). 
While otherwise robust, we did identify one specific failure mode of our model for unseen
predictions, namely that the architecture and training strategy do not allow the model to
learn the typical downregulation of the unseen perturbation target itself (Fig.~\ref{fig:performance_deep_dive}C and Fig.~\ref{fig:performance_deep_dive}D). 
While this error is not inconsequential for understanding the model's abilities, we note that many expression forecasting
methods assign this value to be equal to the ground truth value \citep{kernfeld2023systematic}.

\begin{figure}[tb]
  \centering
  \includegraphics[width=\textwidth]{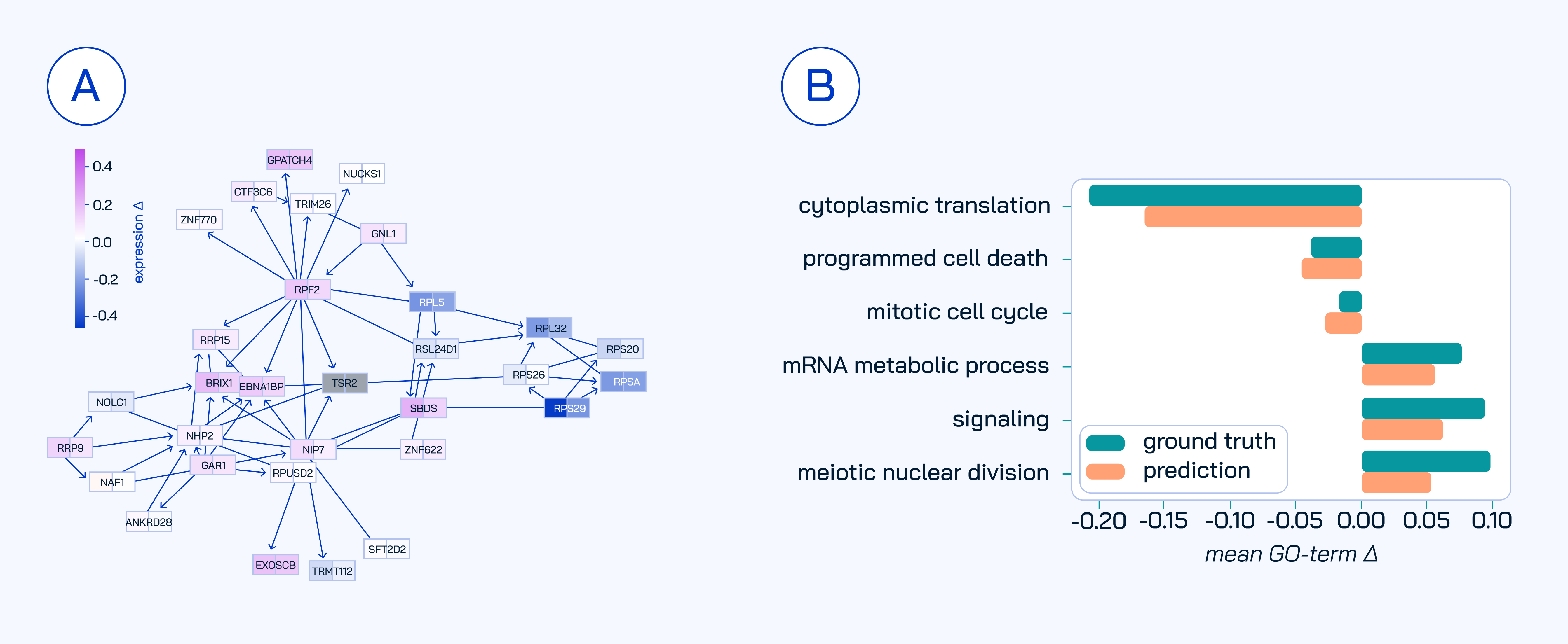}
  \caption{A) Visualization of \modelname~performance in modeling the knockdown of an example held out test gene, TSR2 on  primary and secondary neighbors (up to 30 total, prioritized by weight). Neighbors and edges are from the STRINGdb graph used for modeling. Colors indicate the log fold change vs control cells from min \emph{blue} to max \emph{red} with white centered at 0. The left hand box at each cell shows the ground truth, and the right hand box the prediction. TSR2 is shown in grey as it was not a highly variable gene and thus not predicted; all other nodes with missing values are not shown. B) The expression mean aggregate $\delta$ expression of the 3-most up \& down regulated GO-slim terms (generic set) in the TSR2 knockdown.}
  \label{fig:examples}
\end{figure}

We showcase the performance of the model in predicting the effect of knocking down TSR2, for which defects are associated with Diamond-Blackfan anemia (Fig.~\ref{fig:examples}). \modelname~successfully predicts both local effects of graph neighbors and global function effects (as summarized by GO-slim terms) that are consistent with its role as a ribosome maturation factor. In short, \modelname~makes credible predictions, both in the local graph proximity of a perturbation and 
globally capturing transcriptome wide functional changes.

\section{Discussion}
The past year has brought a reality check to the promise of foundation models in the transcriptomics perturbation domain, as independent benchmarking studies failed to validate the claimed performance of several high-profile models \citep{ihab_benchmark, perturbench, kernfeld2023systematic, wong2025simple, ahlmann2024deep, kedzierska2025zero, csendes2024benchmarking}. In this work, we addressed these concerns through rigorous benchmarking and inclusion of strong baselines, resulting in \modelname, a broadly applicable perturbation model capable of predictions approaching experimental reproducibility levels  on some metrics on unseen perturbations across several datasets. A key factor underpinning our model's success is the effective integration of curated biological databases with large-scale, consistent, and unbiased high-throughput screening data combined with first-class graph modeling. 

Our proposed reusable framework establishes a strong foundation for continued iteration and improvement, and is positioned to benefit further from exciting new developments in the field. For instance, recently released large perturbation single cell datasets \citep{feng2024genome, zhang2025tahoe} could help improve cross-context generalization. Extending our framework towards few-shot or active learning scenarios is another realistic and promising direction to expand beyond the 0-shot cross-cell line setting explored here. Crucially, the field can accelerate progress by continuing to adopt, iteratively refine, and standardize both task definitions and benchmarks. A key next step in improving further \modelname~or other high performing models lies in the inclusion of metrics that explicitly evaluate the conditionality and specificity of perturbation effects in novel contexts.

Although \modelname~exhibits strong performance, it represents just an initial step towards developing  a new generation of models able to accurately model cellular responses to perturbations across diverse biological context. As the community enables further
components, such as compound predictions (e.g. \cite{tong2024transigen, evans2024gsnn}), genomic sequence conditioning \citep{rosen2023uce}, and spatial or intercellular interactions \citep{wen2023cellplm}, the utility of
virtual assays for therapeutic applications will grow. Ultimately, this has the potential to significantly accelerate drug discovery
programs, enable a completely new scope in screening, and open new frontiers for
personalized medicine.

\section{Methods}

\subsection{\modelname~architecture}\label{sec:method}
\modelname~predicts the transcriptional response $\mathbf{y}\in\mathcal{Y}\subset\mathbb{R}^{n}$ given a set of perturbation tokens $P\subset \mathcal{P}\coloneqq\{1,\dots, N\}$ and a basal state representation derived from a control expression profile $\mathbf{x}\in\mathcal{X}\subset\mathbb{R}^{n}$, that has been aligned with the predicted cell with respect to cell line and batch effect. Here, $n\in\mathbb{N}$ denotes the number of experimentally measured genes and $N\in \mathbb{N}$ denotes the total number of observed perturbations in the data. These perturbation tokens (or IDs) are used to select node representations from a biological gene-gene interaction knowledge graph (KG) whose embeddings are integrated with the basal state to produce the perturbed expression profile. 

To combine the information of the basal state and the perturbations, we first learn latent representations of both, i.e., $\mathbf{x} \mapsto\mathbf{s}\in\mathbb{R}^d$ and $P\mapsto \{\mathbf{z}_p\in\mathbb{R}^d : p\in P\}$ for a chosen latent dimension $d\in\mathbb{N}$.
We then combine the information via \emph{latent shift}, where a learned decoder $g_\phi$ predicts the perturbation effect from the given context, i.e., $\hat{\mathbf{y}} = g_\phi(\mathbf{s} + \sum_{p\in P}\mathbf{z}_p)$, using mean squared error (MSE):
\begin{equation*}
       \mathcal{L}(\mathbf{y},\hat{\mathbf{y}})= \frac{1}{n}\mid\mid  \mathbf{y}-\hat{\mathbf{y}}  \mid \mid ^2_2
\end{equation*}
 
This setup naturally integrates with single, double and more general multi-perturbation settings of order $m\coloneqq \vert P\vert$ via the additive and compositional design. More sophisticated combination functions may be used to learn transition functions $\mathbf{s}' = T_\psi(\mathbf{s}, \mathbf{z})$ that allow sequential latent cell state modeling of subsequently applied perturbations, e.g., $\hat{\mathbf{y}} = g_\phi(T_\psi(\dots T_\psi(\mathbf{s},\mathbf{z_{p_1}})\dots,  \mathbf{z_{p_m}}))$.
To obtain $\mathbf{s}$ and $\mathbf{z}_p, p\in P$, we leverage learned encoders, namely the \emph{basal state model} and the \emph{perturbation model}, that are discussed in the following.

\subsubsection{Basal state model}\label{sec:basal}

The basal state encoder is designed to capture intrinsic cellular attributes -- such as cell cycle stage, cell type, and other baseline phenotypic features -- by mapping a control cell’s gene expression profile $\mathbf{x}\in\mathbb{R}^{n}$ to a compact, low-dimensional embedding, $\mathbf{s}=f_\texttt{basal}(\mathbf{x})\in\mathbb{R}^d$. In this work, we employ a multilayer perceptron (MLP) to learn a direct deterministic mapping from high-dimensional input gene expression data to a fixed-size embedding space. The MLP architecture offers a simple and computationally efficient framework for representation learning, while still retaining the capacity to model complex, nonlinear dependencies inherent in gene expression data.

\textbf{Basal information matching \& aggregation:} An important aspect of modeling the basal state is the alignment of the control with the predicted perturbed cell. Beyond choosing this control (randomly sampled) according to the same cell line and dataset or experimental protocol,\footnote{Note that experimental protocols can vary widely between different data sources.} we implement \emph{basal state matching}, where the control cell is selected to closely match the batch metadata of the perturbed sample. As this matching is not unique, we randomly sample one such appropriate control.
We further experiment with \emph{basal state averaging}, where instead of a single control, we compute the average expression profile across all matching controls for a given cell line and/or batch. This produces a more stable estimate of the basal state. Both strategies consistently improved model performance in our experiments.

\textbf{Encoders:} Beyond MLP encoders on raw gene expression profiles, we explored multiple transcriptomics foundation model (FM) embeddings to obtain a latent representation of the basal state. Specifically, we experiment with scGPT, and scVI pre-trained on the Joung dataset \citep{joung2023transcription}. We also include a \emph{no basal state encoder} variant, where the (latent) basal state space is represented directly by the raw expression profile (i.e., $\mathbf{s} \coloneqq \mathbf{x}$). In this configuration, the perturbation decoder learns a \emph{delta} vector from the perturbation embedding, which is added to the control profile: $\mathbf{x}\in\mathbb{R}^{n}$, i.e., $\hat{\mathbf{y}} = \mathbf{x} + g_\phi(\sum_{p\in P}\mathbf{z}_p)$. This resembles the formulation of the general baseline (Section~\ref{sec:eval-general_baseline}) with a trained model predicting the delta instead of a hand-crafted heuristic. Unsurprisingly, this variant shines most in settings with limited data availability for learning robust basal state representations, e.g., perturbation effect prediction in unseen cell lines.

\subsubsection{Perturbation model}\label{sec:pert}
We rely on Graph Neural Networks (GNNs) that leverage biological knowledge graphs (KGs) capturing gene-gene interactions to learn informative embeddings for gene perturbations. The GNN learns a matrix of node embeddings associated with the perturbation tokens $\{1,\dots,N\}\mapsto\mathbf{Z}\in\mathbb{R}^{N \times d}$, with $N\in\mathbb{N}$ the number of perturbations relevant to the investigated task. Each row (or node representation) of this matrix represents the latent encoding of a specific perturbation, i.e., $\mathbf{z}_p\in\mathbb{R}^d$ required for the latent shift is the $p$-th row of $\mathbf{Z}$. More specifically, we first associate each perturbation $p\in\{1,\dots,N\}$ with a randomly initialized input node embedding $\mathbf{h}_p^{0}\in\mathbb{R}^{d_0}, d_0\in\mathbb{N}$ that are consolidated in the input node feature matrix $\mathbf{H}^0\in\mathbb{R}^{N\times d_0}$. During training, those node features are (1) treated as model parameters that are learned using backpropagation and (2) subsequently refined via the message passing of an $L$-layer GNN, i.e., $\mathbf{H}^{\ell} = \texttt{LAYER}_{\theta_\ell}(\mathbf{H}_{\ell-1}), 0\leq\ell\leq L$ and $\mathbf{Z}\coloneqq \mathbf{H}^L$. This allows the model to characterize perturbation effects based on known gene-gene relationships from KGs such as GO~\citep{go}, STRINGdb~\citep{string} or proprietary data sources.

Real-world KGs present inherent imperfections: they often contain noisy or incorrect edges (false positives), suffer from missing connections (false negatives), and may originate from diverse sources offering multiple, sometimes conflicting, perspectives. The effective use of such complex data necessitates GNN architectures specifically chosen to address these challenges.

To this end, we select and adapt two fundamental GNN approaches. First, to handle noisy edges, we employ attention-based models like the Graph Attention Network (GAT) \citep{gat,gat_v2}. The ability of GAT to dynamically (re-)weight neighbor importance provides much needed robustness by effectively down-weighting less credible connections, which is a major difference relative to non-attention-based methods like the simple graph convolution~\citep{sgc} employed in GEARS. Second, to address graph incompleteness and capture long-range dependencies, we utilize Graph Transformers (GTs), specifically Exphormer~\citep{shirzad2023exphormer,shirzad2024even}. Its capacity for attention beyond immediate graph neighbors allows it to potentially model implicit relationships and bridge structural gaps.

Further, it proved crucial for the presented tasks to leverage multiple KGs that offer complementary and reinforcing perspectives on the task-related biology. For this, we explore architectures designed for synergistic learning from diverse sources. This includes extending GAT to \hybridmodel~(allowing for node-level attention weighting of information from different KGs), introducing our provenance-aware Exphormer-MultiGraph (\exphormerMGmodel) variant, and developing \multilayermodel, a multilayer extension of GAT that leverages a supra-adjacency representation to effectively integrate information across multiple biological networks.

In Appendix~\ref{apx:gnns}, we provide rigorous details on the relevant graph representation learning employed for encoding perturbations, the proposed models, and the techniques employed to take advantage of complementary information from multiple KGs. In our experimental setup, the learned embeddings for the perturbed gene(s) from the GNN are extracted and combined with a basal state representation to predict the resulting gene expression profile. This comparative analysis allows us to investigate how different GNN strategies--attention, flexible connectivity, and multi-graph fusion--perform when learning from the complexities of biological knowledge graphs.

\subsubsection{\modelname~model variants \& ablations}\label{sec:ablations}
This work presents multiple modeling options for basal state and perturbation encoding as introduced in Section~\ref{sec:method} so far, which are treated as hyperparameters of the overall \modelname~framework.

With respect to basal matching and averaging, both consistently lead to improved experimental results and are part of the standard configuration of \modelname. While the basal matching mitigates undesired discrepancies between cells subject to different batch effects, control averaging synergizes well with the training and evaluation methodology that aims to predict the mean perturbation effect across samples subject to the same perturbation, e.g., by optimizing mean squared error and calculating metrics aggregated across samples on the perturbation level.

\begin{figure}[tb]
  \centering
  \includegraphics[width=0.8\textwidth]{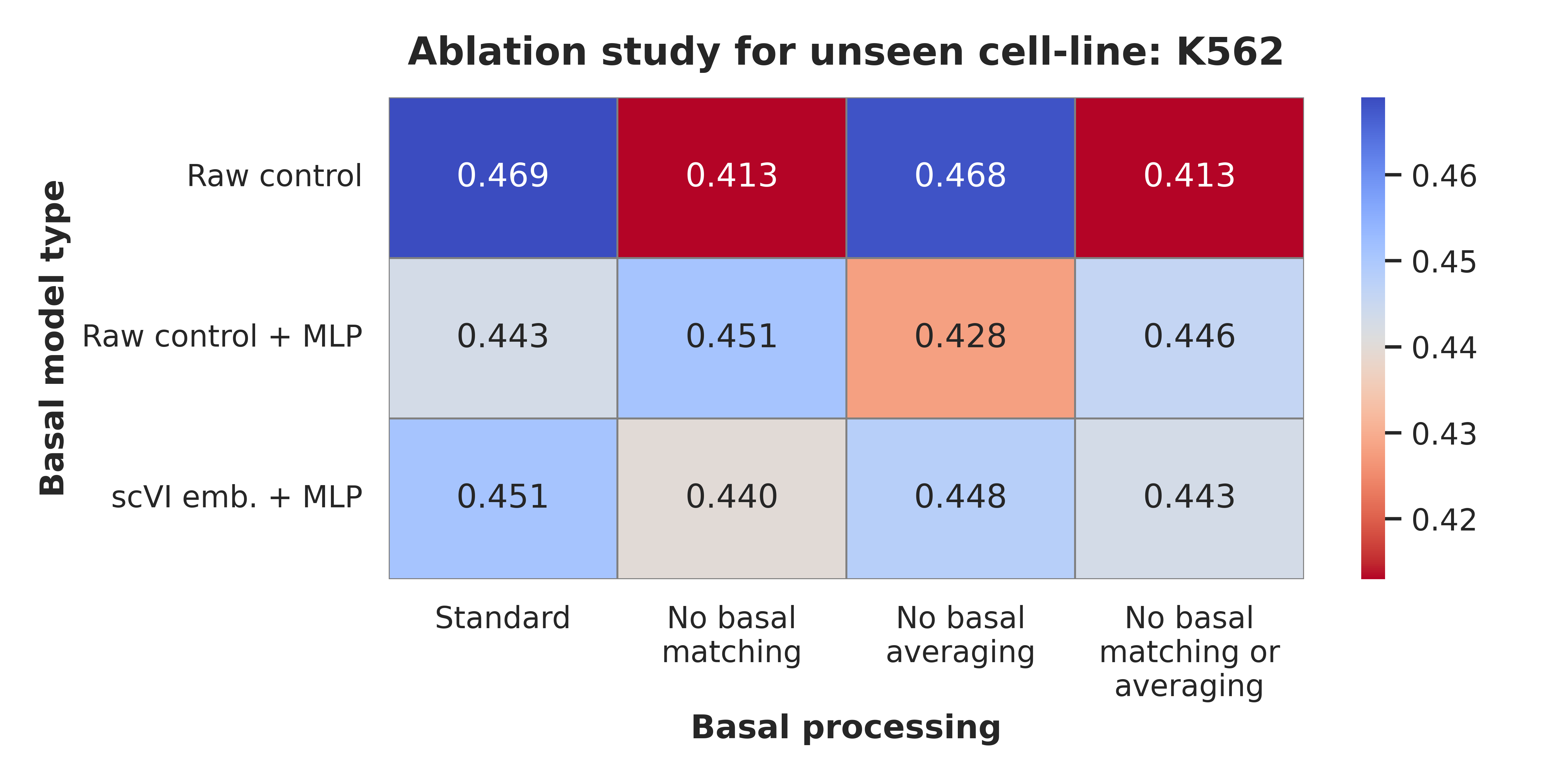}
  \caption{Ablation study of modeling choices for unseen cell line OOD task on K562. We compare the effect of basal state matching and averaging on the different basal modeling choice: no basal state encoder (raw counts) and MLP encoder applied to the raw counts or corresponding FM embedding. As perturbation model, a vanilla GAT using the STRINGdb graph is used. Values are Pearson $\Delta$}
  \label{fig:x-cell-ablation}
\end{figure}

The choice of the basal model type needs more careful investigation, as the effectiveness strongly varies depending on the task and used perturbation model. While \multilayermodel~always performed best with no learned basal state model (i.e., using raw controls), the GAT and Exphormer-based architectures favored the MLP basal encoder.
A unique case is the OOD task of predicting perturbation effects in unseen cell lines (see Fig~\ref{fig:x-cell-ablation}), where using \emph{no basal state encoder} is by far the most effective option. This is a result of the limited availability of diverse cell line data, prohibiting the basal state model from learning generalizable latent representation. We further see in Fig.~\ref{fig:x-cell-ablation}, that learned MLP modeling of the basal state improves when using FM embeddings (here scVI) instead of raw counts, indicating that pretraining on various cell lines can add value. We hope learned basal state models will further improve as more powerful transcriptomic FMs and more data sources with diverse cell line information become available. Another interesting observation is the reduced importance of basal averaging when using scVI embeddings instead of raw counts, suggesting FM embeddings provide more robust encoding of cell line and batch-related information.

\subsection{Training and Evaluation Framework}\label{sec:eval}

\begin{algorithm}
\caption{\modelname~training algorithm}
\begin{algorithmic}[1]
\Require Pert. cells $\mathcal{Y}$, control cells $\mathcal{X}$, biological prior graph $G = (V, E)$ with perturbations $\mathcal{P}\subset V$
\Ensure Minimize MSE loss between predicted and true perturbed cell measurements
\State Initialize input perturbation embeddings $\{\mathbf{h}_v^0\}_{v \in V}$ randomly
\For{each training step}
    \State Sample a batch of perturbed cell profiles: \quad $\{\mathbf{y}_i\}_{i=1}^B \subset \mathcal{Y}$
    \State Sample corresponding control cells from the same experimental batches: \quad $\{\mathbf{x}_i\}_{i=1}^B \subset \mathcal{X}$

    \State Enrich perturbation embeddings using graph prior: \quad $\{\mathbf{z}_v\}_{v \in V}
    \gets \texttt{GNN}(G, \{\mathbf{h}^0_v\}_{v \in V})$
        
    \For{each sample in batch}
        \State Encode control cells into basal latent space: \quad $\mathbf{b}_i \gets \texttt{MLP}_{\text{basal}}(\mathbf{x}_i)$

        \State Retrieve perturbations $P_i\subset \mathcal{P}$ associated to target $\mathbf{y}_i$

        \State Combine control and perturbation embeddings: \quad
        $\hat{\mathbf{z}}_i \gets \texttt{COM}(\mathbf{b}_i, \{\mathbf{z}_p: p\in P_i\})$
        
        \State Decode to predicted perturbed profile: \quad $\hat{\mathbf{y}}_i \gets \texttt{MLP}_{\text{dec}}(\hat{\mathbf{z}}_i)$
        \State Compute loss for each sample: \quad $\mathcal{L}_i \gets \texttt{MSE}(\hat{\mathbf{y}}_i, \mathbf{y}_i)$
    \EndFor
    
    \State Compute total loss over batch: \quad $\mathcal{L} \gets \sum_{i=1}^B \mathcal{L}_i$
    \State Backpropagate and update model parameters
\EndFor

\end{algorithmic}
\end{algorithm}

\subsubsection{Data splits}\label{sec:method-data}

The data was split into train, validation, and test sets via grouping by perturbation
such that distinct sets of unseen perturbations were reserved both the validation and test sets
with a target ratio of (0.5625, 0.1875, and 0.25, for train, validation and test, respectively).
Moreover, for the cross cell type task, the test set was a reserved cell type
with only control cells included during training
with a breakdown into seen and unseen perturbations therein. As an exception, for the doubles task on the Norman dataset pre-defined splits were loaded from
the GEARS setup.

Optimal hyperparameters for each model were selected based on the validation Pearson $\Delta$ metric.
Only metrics on the test set are reported.

\subsubsection{Metric Definitions}\label{sec:eval-metrics}
All metrics are reported as weighted averages, i.e., the mean of the mean across cells subjected to each unique perturbation, unless otherwise specified.

\textbf{Expression value representations and deltas:}
Where not otherwise specified, expression values $\textbf{x}\in\mathcal{X}\cup\mathcal{Y}$, are represented as log1p-transformed and library size-normalized counts (with target library size 4000), i.e., for a raw count $\textbf{x}_{\text{raw}}\in\mathbb{R}^n$, we define
\begin{equation*}
\textbf{x} \coloneqq \log\left(1 + 4000 \cdot \frac{\textbf{x}_{\text{raw}}}{\Vert \textbf{x}_{\text{raw}}\Vert_{1}}\right).
\end{equation*}

The other important representation used is a delta representation, which is centered on batch-matched controls. Specifically, for each perturbed cell expression $\mathbf{y}_i\in\mathcal{Y}$ with  cell line $c$ and batch $b$, the expression is transformed to 
\begin{equation*}
\delta_i \coloneqq \mathbf{y}_i - \overline{\mathbf{x}}_{c,b},
\end{equation*}
where $\overline{x}_{c,b}$ represents the mean expression of control cells $\mathbf{x}\in\mathcal{X}$ with batch $b$ and cell line $c$.

\textbf{Pearson $\Delta$:} Slightly modified from the metric ``Pearson Correlation (Delta Expression)'' from the GEARS manuscript, Pearson $\Delta$ calculates the correlation between predicted and observed log fold change versus batch-matched control mean,
$$
    \texttt{PEARSON}\Delta(p) \coloneqq \texttt{PEARSON}(\hat{\delta}_p, \delta_p),
$$
where $\hat{\delta}_p$ and $\delta_p$ are the batch-matched control centering of the prediction and ground truth, respectively, averaged across replicates of certain perturbation $p\in\mathcal{P}$.\footnote{For simplicity, we define this and following metrics for single perturbations $p\in\mathcal{P}$ but note that analogous formulations are appropriate for multi-perturbations $P\subset\mathcal{P}$.} The results across all predicted perturbation effects are then averaged to obtain an overall performance estimate.

Note that for the GEARS model only, we report the exact ``Pearson Correlation (Delta Expression)'' from the GEARS code base instead. We confirmed that any differences between ``Pearson Correlation (Delta Expression)'' and our ``Pearson $\Delta$'' were much smaller in practice than the differences between models.

\textbf{Retrieval:} We utilize two variants of the retrieval rank metric that score a prediction's similarity to the ground truth not overall, but relative to other perturbations. These metrics are the same as Rank Average from PerturBench, except that they focus on \emph{similarity} with a perfect score of 1, a random score of 0.5, and perfect anti-correlated prediction score of 0: 

\begin{equation*}
    \texttt{RETRIEVAL}(p) \coloneqq \frac{1}{N} \sum_{p\in\mathcal{P}} \text{rank}(\hat{\delta}_p), \quad \text{rank}(\hat{\delta}_p) \coloneqq \frac{1}{N-1} \sum_{\substack{q \in\mathcal{P} \\ q \neq p}} \mathbf{1}_{\{\texttt{PEARSON}(\hat{\delta}_p, \delta_p) \geq \texttt{PEARSON}(\hat{\delta}_p, \delta_q)\}}.
\end{equation*}

For ``normalized'' retrieval the perturbation count $N\coloneqq \vert\mathcal{P}\vert$ matches the original experiment, whereas for ``fast retrieval'' for computational efficiency a seeded random reference set of only 100 perturbations is used, with the addition of the query perturbant $p$ when not in the reference set, so $N \in \{100, 101\}$. Similar to Person $\Delta$, we report averaged performance across all perturbations.

\subsubsection{Non-learned General Baseline}\label{sec:eval-general_baseline}
To establish a performance floor, we implement a non-learned \emph{general baseline} model that predicts expression profiles using mean values observed in the training data. This baseline uses an additive approach that combine the following:
\begin{itemize}
    \item The mean test cell type control expression profile
    \item Either the perturbation-specific mean changes (for seen perturbations) or the global perturbation mean (for unseen perturbations)
    \item When multiple cell lines are present in the training set, we use a weighted average according to the number of samples per cell line
\end{itemize}

Consider a multi-set of training samples $\text{TRAIN}\subset \mathcal{P}_\text{train}\times  \mathcal{C}_\text{train}\times  \mathcal{B}_\text{train}$ consisting of combinations of perturbation(s), cell line and batch effect with a test multi-set \text{TEST} defined analogously. Consider a perturbation $p$ so that $(p, c_p, b_p)\in \text{TEST}$ with cell line $c_p$ and batch effect $b_p$. Implicitly, $(c_p, b_p)$ map to a set of control cell profiles associated to that context.

If there exists $(p, c, b) \in \text{TRAIN}$, we have
$$
    \hat{\mathbf{y}}_{(p, c_p, b_p)} = \overline{\mathbf{x}}_{(c_p, b_p)} + \frac{1}{\vert \{(q,c,b) \in \text{TRAIN}: q=p\}\vert}\sum_{\substack{(q,c,b) \in \text{TRAIN}\\ q=p}} \mathbf{y}_{(p, c, b)} - \overline{\mathbf{x}}_{(c, b)}.
$$

Otherwise, we use the global delta across perturbations observed in $\text{TRAIN}$, i.e.,
$$
    \hat{\mathbf{y}}_{(p, c_p, b_p)} = \overline{\mathbf{x}}_{(c_p, b_p)} + \frac{1}{\vert\text{TRAIN}\vert} \sum_{(q,c,b) \in \text{TRAIN}} \mathbf{y}_{(p, c, b)} - \overline{\mathbf{x}}_{(c, b)}.
$$

For multi-perturbations, this baseline is implemented to initially attempt to use samples where the exact perturbation configuration is present. Otherwise, the perturbation is split into its components and each component is sequentially added to the test control mean according to the above method, adding a local delta estimate if available and resorting to a global delta otherwise.

\subsubsection{Experimental Reproducibility Estimation: Split-Half Validation}\label{sec:eval-exp_acc}
Since PerturbSeq is a destructive assay, we cannot observe the same cell in both perturbed and unperturbed states. This necessitates focusing on distribution means rather than individual cell accuracies. To approximate experimental reproducibility, we use a split-half validation approach:

For each combination of perturbation(s), cell line context and batch, we apply three operations:
\begin{enumerate}
    \item Divide test cells into two roughly equal halves
    \item Calculate mean expression profiles for each half
    \item Measure the agreement between these means using various metrics
\end{enumerate}

To account for the randomness in choosing the half-split, we repeat the experiment across multiple seeded runs and report average performance. This provides a performance benchmark analogous to human-level reproducibility, which is called accuracy in other machine learning domains.

Consider the set of expression profiles $\mathcal{S}\subset\mathcal{Y}$ for a fixed perturbation cell line context and batch $(p, c, b)$ in the test set:

\begin{align*}
    \mathcal{S}' &\subseteq \mathcal{S}: \quad |\mathcal{S}'|\approx |\mathcal{S}| / 2 \\
    \bar{\mathcal{S}}_1 &= \frac{1}{|\mathcal{S}'|} \sum_{\mathbf{y} \in \mathcal{S}'} \mathbf{y} \\
    \bar{\mathcal{S}}_2 &= \frac{1}{|\mathcal{S} \setminus \mathcal{S}'|} \sum_{\mathbf{y} \in \mathcal{S} \setminus \mathcal{S}'} \mathbf{y}.
\end{align*}

We then report
\begin{equation*}
    \texttt{REPRODUCE}(p, c, b) = \texttt{METRIC}(\bar{\mathcal{S}}_1, \bar{\mathcal{S}}_2),
\end{equation*}
where $\texttt{METRIC}$ represents any of our evaluation metrics, e.g., Pearson $\Delta$, \texttt{RETRIEVAL}, or \texttt{MSE}. Theoretically, the experimental reproducibility is not guaranteed to establish an upper bound for performance of all models at test time because it operates on a different test set (only using half for prediction and testing, respectively). However, it empirically proves to be very close to an upper bound of any other method explored here, with only one exception when models slightly outperformed it for prediction unseen perturbation effects on K562.

\subsection{Data}\label{sec:data}
\subsubsection{PerturbSeq data sources}
We demonstrate the efficacy of our approach across a range of datasets, including CRISPRi (gene knockdown) of ~2k essential genes in K562 and RPE1 cell lines from~\citet{replogle} (also used in GEARS~\citep{gears}) and similarly designed CRISPRi experiments in Jurkat and HEPG2 cell lines from~\citet{nadig}. Further the Norman \citep{norman2019exploring} dataset with %
94 unique single and 110 unique double CRISPRa (gene over expression) perturbations respectively in the K562 cell line. 

\subsubsection{Graphs - sourcing and processing}
The graphs used as inductive bias in this work can be classified into two main categories, 1. Curated publicly available biological knowledge, and 2. Large-scale perturbation screens.

The curated graphs from category 1 include the GO graph, first used by GEARS, which is constructed by assigning edges between nodes that have a high Jaccard Index in their GO-terms \citep{go}; the (Search Tool for the Retrieval of Interacting Genes) STRINGdb graph \citep{string}; and Reactome \citep{reactome}. 

Category 2 graphs are generated from large-scale perturbation screens including DepMap \citep{depmap} and Perturb-seq \citep{perturb-seq}. These are extensive datasets linking genetic perturbation to either morphological or transcriptomic outcomes, which can offer particularly crucial insights into cellular responses to stimuli. To translate these experimental screens into graphs we utilize derived embeddings to represent the genes and cell lines in a high-dimensional space, allowing for the analysis of relationships and identification of dependencies.

To curate these graphs we first compute the pairwise similarity score between all combinations of genes. This means that, for each pair of genes $(g_i, g_j)$, we compute the cosine similarity between their (aggregated) embeddings $\mathbf{x}_{g_i}$ and $\mathbf{x}_{g_j}$. Cosine similarity is computed as follows:

\begin{align*}
\text{cosine similarity}(\mathbf{x}_{g_i}, \mathbf{x}_{g_j}) &= \frac{\mathbf{x}_{g_i} \cdot \mathbf{x}_{g_j}}{\|\mathbf{x}_{g_i}\|\|\mathbf{x}_{g_j}\|} \\
&= \frac{\sum_{k=1}^{n} \mathbf{x}_{g_ik} \mathbf{x}_{g_jk}}{\sqrt{\sum_{k=1}^{n} \mathbf{x}_{g_ik}^2} \cdot \sqrt{\sum_{k=1}^{n} \mathbf{x}_{g_jk}^2}}
\end{align*}

where
\begin{itemize}
    \item $\mathbf{x}_{g_i} \cdot \mathbf{x}_{g_j}$ represents the dot product of the vectors
    \item $\|\mathbf{x}_{g_i}\|$ and $\|\mathbf{x}_{g_i}\|$ represent the Euclidean norms (magnitudes) of vectors $\mathbf{x}_{g_i}$ and $\mathbf{x}_{g_j}$ respectively
    \item $\mathbf{x}_{g_ik}$ and $\mathbf{x}_{g_jk}$ are the individual components of vectors $\mathbf{x}_{g_ik}$ and $\mathbf{x}_{g_jk}$.
\end{itemize}

These cosine similarities are converted to their absolute values because the difference between highly cosine negative and highly cosine positive does not translate literally to the signed weight of the edge in the graph.

We additionally leverage proprietary data from internal genome-wide perturbation screens where we measure the similarity of perturbation effect using both microscopy imaging and transcriptomics in various cell types. 

Filtering configurations were optimized empirically. We found that the most performant configuration involved selecting for the top 1\% of edges by (absolute) weight for screen-based graphs. For all other graph types we (additionally) filtered for no more than 20 incoming nodes by target. Edge direction was assigned arbitrarily for undirected edges.

\subsubsection{Data Understanding}
The following subsection contains additional methods related to specific analyses.

\textbf{Pharos knowledge rank:} The Pharos initiative consolidates a variety of statistics relating to how researched
and well known specific genes are \citep{sheils2021pharos}. Starting from this we ranked knowledge levels as the
mean of the rank of the Pharos Pubmed score, and the rank of the Pharos negative log novelty score
to create a single Pharos knowledge rank. We used this rank to break down and compare to the performance of
models, and understand potential bias. 
The ``knowledge levels'' 0, 1, 2, and 3 correspond to the following bins of the Pharos knowledge rank:
\begin{itemize}
    \item knowledge level 0 (least characterized): 0 - 0.125
    \item knowledge level 1: 0.125 - 0.25
    \item knowledge level 2: 0.25 - 0.5
    \item knowledge level 3 (most characterized): 0.5 - 1.
\end{itemize}

\textbf{Within vs across:} In investigating the correlations between controls and mean baselines, we compared ``within'' context correlation to ``across''. 
Generally prior to calculating either, all examples were first split into two mutually exclusive halves, A and B; 
``within'' context correlation is a comparison of A vs B, in each context, while
``across'' context is a comparison of an arbitrary half of one context to another context. The only exception is ``across'' batch controls in Fig.~\ref{fig:dataanalysis}A, for which, to make a conservative
estimate of across batch variance, full batches were aggregated without splitting.
For batch comparison, individual control cells were split and aggregated; for
the mean baselines, the $\delta$ of perturbant replicate cells was pre-aggregated,
and then split (such that the halves had non-overlapping perturbations). 

\textbf{Functional enrichment:} To achieve a descriptive biological summary of
the actual gene expression changes in the mean baseline, we first calculated 
a meta mean baseline (mean of all cell types and datasets, using the intersect 
of provided expressed genes), and defined 
up/down-regulated genes as having a remaining $\delta > 0.05$ or
$\delta < -0.05$, respectively. We then ran functional enrichment testing separately on each
on these sets (versus the background of all genes in the dataset intersect) using the GOATOOLS package \citep{klopfenstein2018goatools}.

\textbf{Metric selection via retrieval:} To avoid confusion and
distraction caused by reporting many similarly performing, or perhaps
slightly contradicting metrics, we first performed an empirical
``test of the test metrics''. We adapted the evaluation method pioneered by
\citet{szalata2024benchmark} to work for our data and task. In short,
this method uses cross-context retrieval of a perturbation as a way
to judge whether a representation and metric together allow the 
retention and comparison of details necessary to distinguish perturbations. In our case, we modified the method to work on 
non-aggregated single cell expression profiles (as this is 
the input and output of our model), and ran retrieval across the essential perturbation set on
core cell types (K562, RPE1, HEPG2 and Jurkat). For each retrieval
calculation, instead of aggregating we first randomly sampled one cell of each perturbation. We ran three replicates on each of the cell $\times$ cell
pairings for $n = 3  \cdot 6 = 18$ total estimates per perturbation.
We report the 0.9 quantile, to focus on active perturbants; however,
similar patterns were observed at other thresholds. 

Note that the choice to focus on single cells excluded use of the 
representation selected by \citet{szalata2024benchmark}, namely
the signed p-value. To de-risk this we ran a preliminary analysis
on the exact setup as \citet{szalata2024benchmark} and note that
we were only able to reproduce their results when making choices
that would have limited the extensibility of our data and training
setup; in particular, we found the high performance of the 
signed p-value to rely
on performing a global fit (and thus using a global estimate for
gene-wise variance) across all contexts for 
determining differential expression. 

We focused our selection of representations and metrics especially
common in the perturbational transcriptomics literature, but do 
acknowledge the current omission of count-based representation and metrics.

\section{Data availability}
All data used here is already publicly available; with the exception of 
the PxMap and TxMap graphs, which are Recursion proprietary assets.
Pre-processed and model ready data is automatically downloaded via the code below.

\section{Code availability}
The code to reproduce results with public data is available at \url{https://github.com/valence-labs/TxPert}.

\section{Authors contributions}
Using the CRediT system: Conceptualization: FW, WT, CM \& AD; Data curation: CM \& SW; Formal analysis: AD; Investigation: FW, WT, HS, CM, CE, IB, CR, \& AD; Methodology: FW, WT, CM, HS, CE, IB, \& CR. 
Project administration: FW \& AD; Resources: BE; Software: FW, WT, CM, HS, CE, IB, CR, LH, YEM, \& AD; Supervision: FW, JD, MF, BE, EN, \& AD; Validation: FW, WT, CE, \& AD; 
Visualization: FW, WT, CM, HS, SW, IB, LH \& AD; Writing -- original draft: FW, WT, CM, HS, CE \& AD; Writing -- review and editing: FW, WT, CM, HS, SW, IB, LH, MF, EN, \& AD.

\section{Acknowledgments}
We thank Jonathan Hsu and Humza Salam for their assistance with the figure design.

\bibliography{refs}
\bibliographystyle{valence}

\newpage

\appendix

\appendix

\begin{center}
  \Large \textbf{\rawmodelname~Appendix : Leveraging Biochemical Relationships for Out-of-Distribution Transcriptomic Perturbation Prediction}
\end{center}

\vspace{1em}  %

\section{GNN perturbation encoder model details}\label{apx:gnns}

\textbf{General graph network structure.}
We consider a simple weighted graph \( G = (V, E, w) \) with a set of $N^\star\in\mathbb{N}$ nodes (or genes/perturbations) \( V=\{v_1, \dots, v_{N^\star}\} \) and edges \( E\subset V\times V \) representing gene-gene interactions. Note that $N \leq N^\star$ is potentially larger than number of observed perturbations $N$. This means our models may reason along gene-gene interactions and pathways containing genes that are not necessarily observed as a perturbation in the data. We write \( v \sim u \) if \( \{v,u\} \in E \). We define the neighborhood of a node as 
$
    N(v)=\{u \in E: v \sim u\}.
$
Additionally, an edge weight \( w: E \rightarrow \mathbb{R} \) characterizes the strength of the interaction. The derivation of the edge weight differs significantly between different graphs and data sources.
We use cosine similarity of embeddings from both internal and publicly available foundation models for graphs derived from the large-scale perturbation screens (\( w: E \rightarrow [-1,1] \)), while literature-based graphs like STRINGdb and GO (\( w: E\rightarrow [0,1] \)) factor in information from experiments \emph{and} the literature.
The graph structure is maintained in the adjacency matrix \( \mathbf{A}\in \mathbb{R}^{N^\star \times N^\star} \), where
$$
    A_{ij}\coloneqq
    \begin{cases} 
        w(e) & \text{if } e=\{v_i, v_j\} \in E, \\ 
        0 & \text{otherwise}.
    \end{cases}
$$

\textbf{Node features:}
The input node features for each perturbation are randomly initialized using the Kaiming distribution. Specifically, the vectors \( \mathbf{h}_v^0 \in \mathbb{R}^{d_0} \) are drawn from this distribution before being refined via backpropagation. The Kaiming distribution was chosen to help reduce exploding and vanishing gradients.
We concatenate the input node features across perturbations to form the input node feature matrix \( \mathbf{H}^0\in\mathbb{R}^{N^\star\times d_0} \).

\textbf{Graph Attention Network:}
Our base GNN architecture is based on an architecture from~\citet{gat_v2}, which is a variation of the graph attention network~\citep[GAT;][]{gat} that offers improved theoretic expressive power. Overall, we use $L\in\mathbb{N}$ layers where
$$
    \mathbf{H}^\ell = f(\mathbf{H}^{\ell-1},\mathbf{A},\mathbf{W}^\ell) \quad \text{and} \quad \mathbf{H}^L = F(\mathbf{H}^0) = f^L \circ f^{L-1} \circ \dots \circ f^1 (\mathbf{H}^0).
$$
This model uses an attention mechanism to aggregate information during message-passing (MP). For every edge, we learn an attention score that characterizes its importance that is then normalized at every node across incoming edges. On the node level, the features of a node $v\in V$ are updated as follows:
$$
    \mathbf{h}_v^\ell = \sigma \left(\sum_{u\in N(v)} \bar{a}_{uv} \mathbf{W}^\ell \mathbf{h}_u^{\ell-1} \right).
$$

To derive the attention scores $\bar{a}_{uv} = \texttt{ATT}_{\theta, \mathbf{W}}(\mathbf{h}_u, \mathbf{h}_v)$, we first calculate
$
    a_{uv} = \theta^\top \sigma \left(\left[\mathbf{W} \mathbf{h}_u \Vert \mathbf{W} \mathbf{h}_v \right] \right),
$
followed by
$$
    \bar{a}_{uv} \coloneqq \texttt{Softmax}\left( a_{uv} \vert N(v)\right) \coloneqq \frac{\exp(a_{uv})}{\sum_{w\in N(v)} \exp(a_{wv})}.
$$

\textbf{Hybrid GNN:} We further propose a \hybridmodel~that learns perturbation embeddings by combining the information from multiple GNN channels (i.e., different GNN types and/or graphs) using a data-driven node level attention mechanism. Related architectures have been used in traditional graph representation learning and combinatorial optimization~\citep{hybrid,wenkel2024towards}.

For a set of GNN channels $\mathcal{F} = \{F_1,\dots,F_K\}$, we first aggregate the outputs each channel $\{H^F:F\in \mathcal{F}\}$. 
We then calculate a node-to-channel attention score $s_F: V \rightarrow \mathbb{R}$,
$$
    s_F(v)\coloneqq \vartheta^\top \sigma\left([\mathbf{W} \mathbf{h}_v^0 \Vert \mathbf{W} \mathbf{h}_v^F] \right).
$$
Similar to a conventional GAT, we normalize the attention scores using the Softmax function, not across nodes but rather across channels $\bar{s}_F(v) \coloneqq \texttt{Softmax}\left(s_F(v) \vert \mathcal{F}\right)$. We obtain the aggregated perturbation embeddings by weighting channels according to their attention score,
$$
    \mathbf{h}_v =\sum_{F\in\mathcal{F}} \bar{\mathbf{s}}_F(v) \mathbf{W} \mathbf{h}_v^F.
$$
Finally, we apply a multi-layer perceptron to the perturbation embeddings, $\mathbf{h}_v^\prime = \texttt{MLP}(\mathbf{h}_v)$.

\hybridmodel~is particularly suited for the investigated OOD tasks as it provides a flexible framework for combining information from multiple graphs into informative node representations that here featurize the genetic perturbation effects. Through the specialized attention mechanism, the model can learn to rely on different graph priors depending on the node/gene in question.

\textbf{Exphormer:}
Graph Transformers have shown strong performance recently, especially for modeling long-range dependencies \citep{dwivedi2020generalization,rampavsek2022recipe,shirzad2023exphormer}.
Standard Graph Transformers often use a dense self-attention mechanism, similar to the original Transformer \citep{vaswani2017attention}.
In this setup, every token attends to every other token.
This leads to computational and memory costs that grow quadratically with the number of tokens ($n$) \citep{dwivedi2020generalization,ying2021transformers,kreuzer2021rethinking}.
Such quadratic costs can be too high when working with the large number of genes involved in our task.
To handle this computational challenge, various sparse and efficient Transformer architectures have been proposed \citep{tay2022efficient}.

One promising direction in graph learning involves using expander graphs, a concept studied extensively in the community \citep{shirzad2023exphormer,deac2022expander,shirzad2024even,wilson2024cayley,shirzad2024theory}.
In this work, we adapt a variant of the Exphormer model to function as a perturbation modeler.
Exphormer builds a sparse attention pattern by combining the original graph's edges with edges from an expander graph.
This combined structure preserves the graph's inductive bias while improving information flow through shortcut paths and offering stronger theoretical guarantees \citep{di2023does}.
The specific Exphormer variant we adopt is partially based on the follow-up work, called Spexphormer, which incorporates edge type bias into the model \citep{shirzad2024even}.

This model uses a multi-head attention mechanism on a sparse attention pattern composed of the graph structure and an expander graph. The sparse attention pattern is denoted by $\mathcal{R}$, and $N_\mathcal{R}(i)$ denotes the set of neighbors for node $i$ according to pattern $\mathcal{R}$. Let $\mathbf{H} = [\mathbf{h}_1, \dots, \mathbf{h}_{N^\star}]^\top \in \mathbb{R}^{N^\star \times d}$ be the matrix of node features from the previous layer, or initial node features. The model computes the updated feature vector for node $i$ using $t$ attention heads as follows:
\[
\texttt{ATTN}_\mathcal{R}(\mathbf{\mathbf{H}})_{i} \coloneqq \mathbf{h}_i + \sum_{j=1}^t \mathbf{V}_i^j \sigma\left( (\mathbf{E}_i^j \odot \mathbf{K}_i^j)^T \mathbf{Q}_i^j + \mathbf{B}_i^j \right)
\]
where:
\begin{itemize}
    \item $\mathbf{Q}_i^j = \mathbf{W}_Q^j \mathbf{h}_i$ is the query vector derived from node $i$'s features using projection matrix $\mathbf{W}_Q^j$
    \item $\mathbf{K}_i^j = \mathbf{W}_K^j \mathbf{H}^\top_{N_\mathcal{R}(i)}$ and $\mathbf{V}_i^j = \mathbf{W}_V^j \mathbf{H}^\top_{N_\mathcal{R}(i)}$ are the key and value matrices, respectively. They are computed by applying linear projections $\mathbf{W}_K^j$ and $\mathbf{W}_V^j$ to the features of the neighboring nodes $\mathbf{H}_{N_\mathcal{R}(i)} = [\mathbf{h}_u]^\top_{u \in N_\mathcal{R}(i)}$
    \item Edge features are incorporated via $\mathbf{E}_i^j = \mathbf{W}_E^j \mathcal{E}_{N_\mathcal{R}(i)}$ and $\mathbf{B}_i^j = \mathbf{W}_B^j \mathcal{E}_{N_\mathcal{R}(i)}$. Here, $\mathcal{E}_{N_\mathcal{R}(i)} \in \mathbb{R}^{d_E \times |N_\mathcal{R}(i)|}$ contains the features for edges connecting neighbors in $N_\mathcal{R}(i)$ to node $i$. $\mathbf{W}_E^j$ and $\mathbf{W}_B^j$ are learnable projection matrices. The symbol $\odot$ denotes element-wise multiplication, allowing edge features $\mathbf{E}_i^j$ to modulate the keys $\mathbf{K}_i^j$. $\mathbf{B}_i^j$ acts as an edge type attention bias term
    \item $\sigma$ represents the softmax function applied over the neighbors in $N_\mathcal{R}(i)$, calculating the attention weights. The expression $\mathbf{V}_i^j \sigma(\dots)$ computes the weighted sum of value vectors based on these attention weights for head $j$.
\end{itemize}
The edge features $\mathcal{E}$ themselves are typically learnable embeddings specific to each edge type (e.g., original graph edges vs. expander edges). This mechanism allows the model to differentiate between information flowing along different types of connections within the attention pattern $\mathcal{R}$. Similar to conventional Transformer models, each attention layer is followed by node-wise feed-forward-network with one hidden layer to create the output node features.

Building upon the Exphormer variant from \citet{shirzad2024even} (which introduced edge type bias), we introduce further modifications addressing how different edge sources are combined.
The original Exphormer design incorporated edges from both the input graph and an expander graph into its attention mechanism, using learnable embeddings to distinguish between these edge types.
However, in that setup, an edge could potentially appear in both sources, leading to duplicate connections within the attention pattern.
To address this potential duplication, we propose an enhanced approach that not only avoids edge duplication but also improves the Exphormer architecture's efficiency and enables the simultaneous integration of multiple knowledge graphs within the model framework.

Specifically, we take the union of all edges across the involved knowledge graphs and the expander graph.
Rather than duplicating edges that appear in multiple sources, we encode their provenance using multi-hot edge features.
These features indicate which source graphs (including the expander graph) each edge belongs to.
More mathematically, assume we have $k$ source graphs (e.g., knowledge graphs or an expander graph) with a shared set of nodes $V$ and edge sets $E_1, E_2, \dots, E_k$.
We construct a unified graph with the node set $V$ and an edge set defined by the union $E = E_1 \cup E_2 \cup \dots \cup E_k$.
For each edge $e \in E$, we define a $k$-dimensional multi-hot feature vector $\mathcal{E}_e \in \{0, 1\}^k$. The $j$-th component of this vector, denoted as $(\mathcal{E}_e)_j$, indicates membership in the $j$-th source graph:
\begin{equation*}
(\mathcal{E}_e)_j =
\begin{cases}
  1, & \text{if edge } e \in E_j \\
  0, & \text{otherwise}.
\end{cases}
\end{equation*}

This design allows the Exphormer model variant to use multiple knowledge graphs simultaneously.
The computational cost relates to the total number of unique edges in the union ($|E| = |E_1 \cup \dots \cup E_k|$), which can be significantly less than the sum of edges across all graphs ($\sum |E_i|$) if there is substantial overlap, thus offering computational efficiency compared to processing graphs independently.
Through these edge features (encoding provenance), the model can effectively combine structural information from various knowledge graphs with the enhanced information propagation properties provided by the expander graph component.
For clarity in our discussions, we will use ``Exphormer'' to refer to the model using a single primary knowledge graph alongside the expander graph, and ``\exphormerMGmodel'' (Multi-Graph) for the version that integrates multiple knowledge graphs (and the expander graph) using our proposed union and multi-hot feature approach.

\textbf{MultiLayer graphs:} 
Inspired by taking a holistic view for understanding proximity and influence in a network (\cite{zangari2024link, yun2021neo}), we consider a MultiLayer modified graph attention mechanism for learning the node embeddings in the perturbation model. Existing methods tend to learn layer-specific knowledge which is bias towards the edge types in the network, and more specifically, what is deemed an edge. Therefore we introduce \multilayermodel~for learning the perturbation embeddings of the genes.

Given a set $\mathcal{V}$ of $\textit{n}$ entities and a set $\mathcal{L} = \{L_1, \cdots , L_\ell \}$ of layers, we denote a multilayer network by $\mathcal{G_L} = \langle V_\mathcal{L}, E_\mathcal{L}, \mathcal{X},\mathcal{V}, \mathcal{L} \rangle$, where $V_\mathcal{L} \subseteq \mathcal{V} \times L$ is the set of all entity occurrences, or nodes, in $\mathcal{L}$. $V_l$ is the set of nodes in layer $l$ $(l \in \mathcal{L})$. $E_{\mathcal{L}}$ is the set of edges between nodes belonging to the same layer, and $E_l \subseteq V_l \times V_l$ is the set of edges in layer $l$. Each entity has a node in at least one layer, hence $\mathcal{V} = \bigcup_{l=1 .. \ell} V_l$ and inter-layer edges exist between each node in a layer and its counterpart in a different layer.

A multilayer network can be represented by a set of adjacency matrices $\mathcal{A} = \{\mathbf{A}_1, \cdots , \mathbf{A}_{\ell} \}$, with $\mathbf{A}_l \in \mathbb{R}^{n_l \times n_l} (l \in L)$, where $n_l = |V_l|$. Entities can be associated with external information or attributes stored in layer-specific matrices $\mathcal{X} = \{\mathbf{X}_1, \cdots \mathbf{X}_{\ell}\}$, where $\mathbf{X}_l$ is the attribute matrix for layer $l$. If no attributes are given for nodes in a layer, we randomly initialize them.

To work in the multilayer setting we adapt the GATv2 architecture from~\citet{gat_v2}. Our approach is inspired by the work of~\citet{zangari2024link} and first transforms the multilayer network into a supra-adjacency representation, which encodes both intra-layer and inter-layer connections in a unified graph structure. This allows us to leverage the standard GATv2 mechanism while still capturing the multilayer representation of the data.

The node features for each gene representation are learned through an embedding layer. Specifically, our model represents each node in each layer with a learnable embedding vector and constructs a supra-adjacency matrix that connects nodes both within and across layers. Formally, given $\ell$ distinct layers, we define the supra-adjacency matrix $\mathbf{A}_S$ as:

\begin{equation*}
\mathbf{A}_S =
\begin{pmatrix}
\mathbf{A}_1 & \mathbf{I}_{1,2} & \cdots & \mathbf{I}_{1,\ell} \\
\mathbf{I}_{2,1} & \mathbf{A}_2 & \cdots & \mathbf{I}_{2,\ell} \\
\vdots & \vdots & \ddots & \vdots \\
\mathbf{I}_{\ell,1} & \mathbf{I}_{\ell,2} & \cdots & \mathbf{A}_{\ell}
\end{pmatrix}
\end{equation*}

where $\mathbf{A}_i$ is the adjacency matrix for layer $i$, and $\mathbf{I}_{i,j}$ is a matrix defining the inter-layer connections between layers $i$ and $j$, typically identity matrices that connect the same entity across different layers.

Using this representation, we apply the GATv2 mechanism to learn node embeddings. For a node $v$ in the supra-adjacency graph, its feature update at layer $k$ of the neural network is:

\begin{equation*}
\mathbf{h}_v^{(k)} = \texttt{AGG}_{q=1}^Q \left( \sum_{u \in \mathcal{N}(v)} \alpha_{(v,u)}^{(q,k)} \mathbf{W}^{(q,k)} \mathbf{h}_u^{(k-1)} \right)
\end{equation*}

where $\mathcal{N}(v)$ represents all neighbors of node $v$ in the supra-adjacency graph (including both intra-layer and inter-layer connections), $\alpha_{(v,u)}^{(q,k)}$ is the attention coefficient between nodes $v$ and $u$ for the $q$-th attention head at the $k$-th layer, $\mathbf{W}^{(q,k)}$ is the learned weight matrix, and $\mathbf{h}_u^{(k-1)}$ is the feature vector of node $u$ from the previous layer.

The attention coefficients are computed as:

\begin{equation*}
\alpha_{(v,u)}^{(q,k)} = \frac{\exp\left(\texttt{LeakyReLU}\left(\mathbf{a}^{(q,k)T}[\mathbf{W}^{(q,k)}\mathbf{h}_v^{(k-1)} \Vert \mathbf{W}^{(q,k)}\mathbf{h}_u^{(k-1)}]\right)\right)}{\sum_{w \in \mathcal{N}(v)} \exp\left(\texttt{LeakyReLU}\left(\mathbf{a}^{(q,k)T}[\mathbf{W}^{(q,k)}\mathbf{h}_v^{(k-1)} \Vert \mathbf{W}^{(q,k)}\mathbf{h}_w^{(k-1)}]\right)\right)}
\end{equation*}

where $\mathbf{a}^{(q,k)}$ is the attention vector for the $q$-th head at the $k$-th layer, and $\Vert$ represents concatenation.

For the aggregation function across attention heads, we implement two strategies:

\begin{equation*}
\texttt{AGG}_{q=1}^Q =
\begin{cases}
\texttt{CONCAT}_{q=1}^Q(\mathbf{h}_v^{(q,k)}), & \text{if aggregation = `concat'} \\
\frac{1}{Q}\sum_{q=1}^Q \mathbf{h}_v^{(q,k)}, & \text{if aggregation = `avg'}
\end{cases}
\end{equation*}

After propagating information through multiple GATv2 layers, we extract the final layer embeddings, which capture both local and global structural information across the multilayer network. These embeddings serve as comprehensive representations of each gene's perturbation effect in the context of multiple biological interaction networks.

\multilayermodel~provides a perturbation modeling approach that is robust to heterogeneous network structures and can effectively capture cross-layer dependencies between genes. By simultaneously considering multiple biological interaction networks, this model learns comprehensive gene embeddings that reflect both pathway-specific and global effects of perturbations. This enables more accurate predictions for unseen single perturbations, where the model can leverage information from related genes across different layers. Moreover, for double perturbations, the model effectively captures the complex non-linear effects that emerge from simultaneous modifications of multiple genes, even when these specific combinations were not observed during training. 

\section{Supplemental Figures and Tables}
We report supporting figures and tables here that provide additional context for the results presented in the main text.

\begin{figure}[h]
  \centering
  \includegraphics[width=0.7\textwidth]{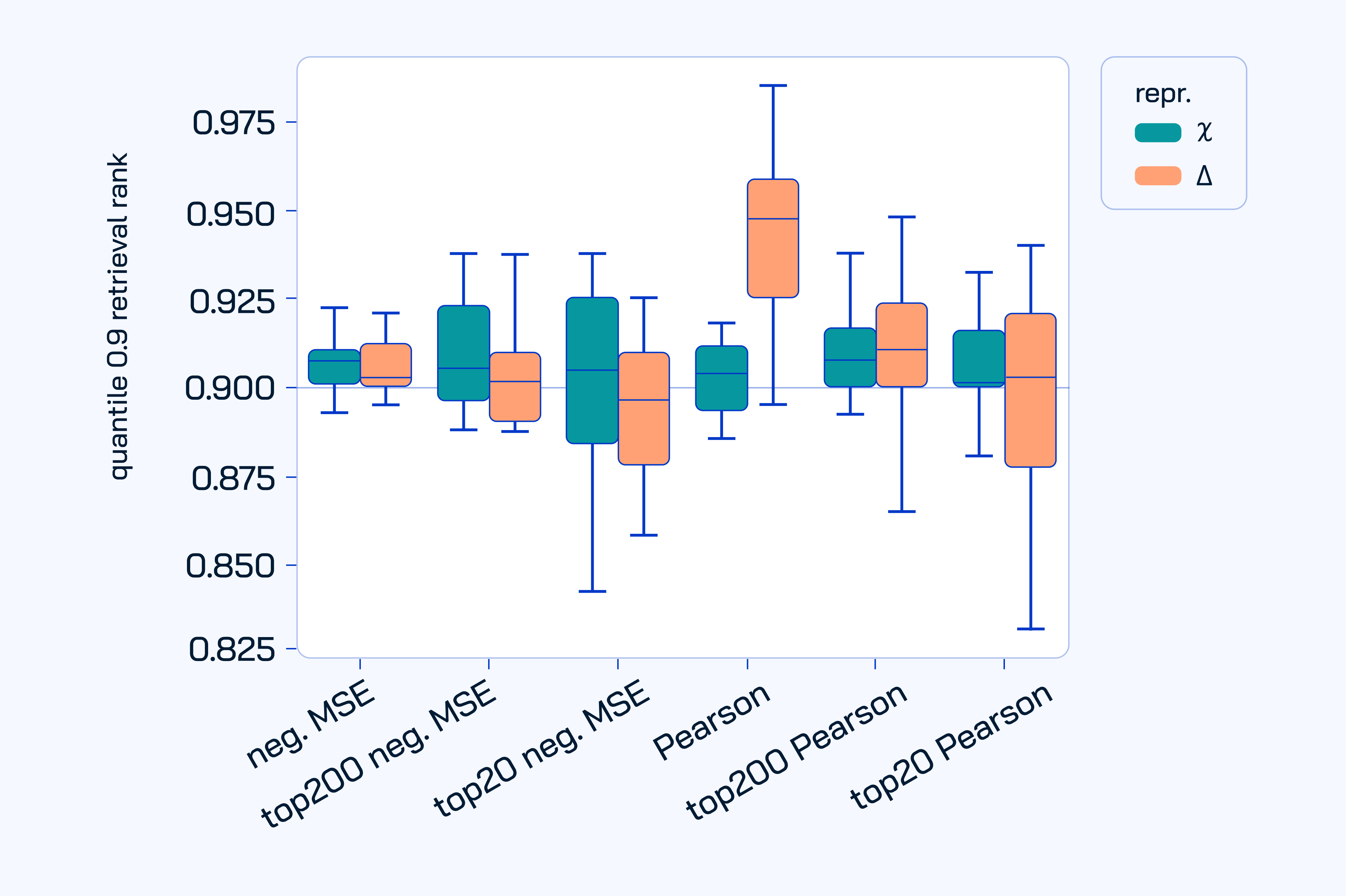}

  \caption{Normalized retrieval between true perturbant replicates in different biological contexts. Retrieval is calculated based on the indicated expression representations and similarity metric. The ``top'' categories representing calculating the metric on a subset of 20 or 200 observed genes, specifically the most significantly differentially regulated under each perturbation according to a Wilcoxon test, as implemented by Scanpy. The plotted value is the 0.9 quantile (across all unique perturbants), where expected random performance is 0.9, indicated by the dashed line.}
  \label{fig:no_top_DE}
\end{figure}

\begin{figure}[h]
  \centering
    \includegraphics[width=\textwidth]{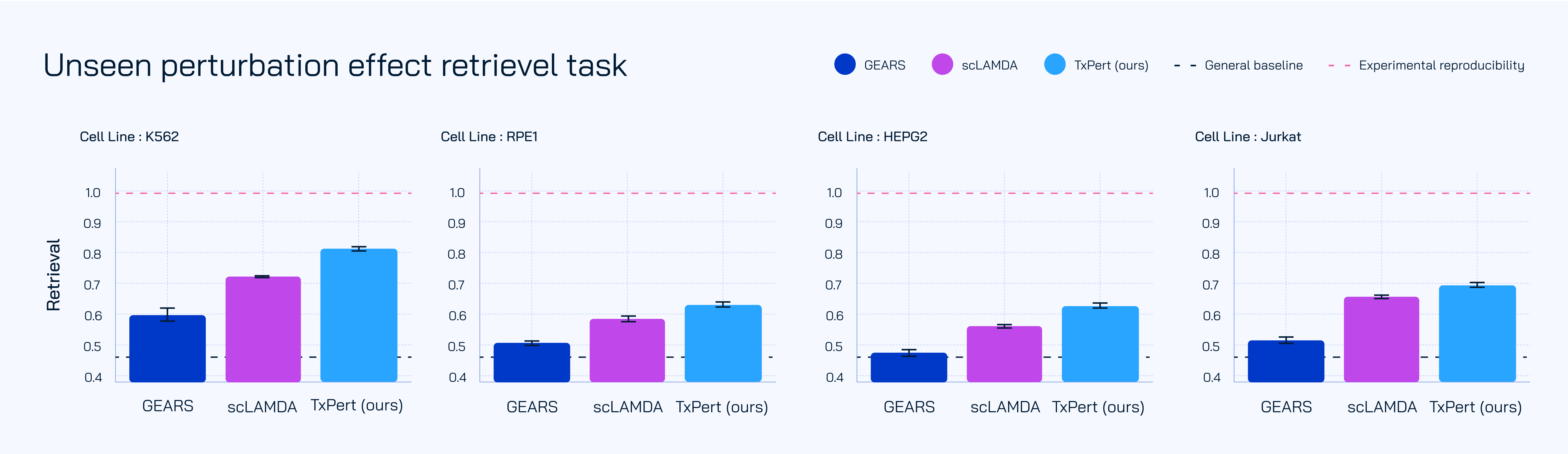}

  \caption{Performance of \modelname~compared to GEARS and scLAMBDA on retrieving unseen single perturbations using predicted perturbation effect within a known cell type. Horizontal bars indicate general baseline, a batch-informed model (capturing potential confounding), and experimental reproducibility (see Section~\ref{sec:method}).}
  \label{fig:sota_unseen_retrieval}
\end{figure}

\begin{figure}[h]
  \centering
    \includegraphics[width=\textwidth]{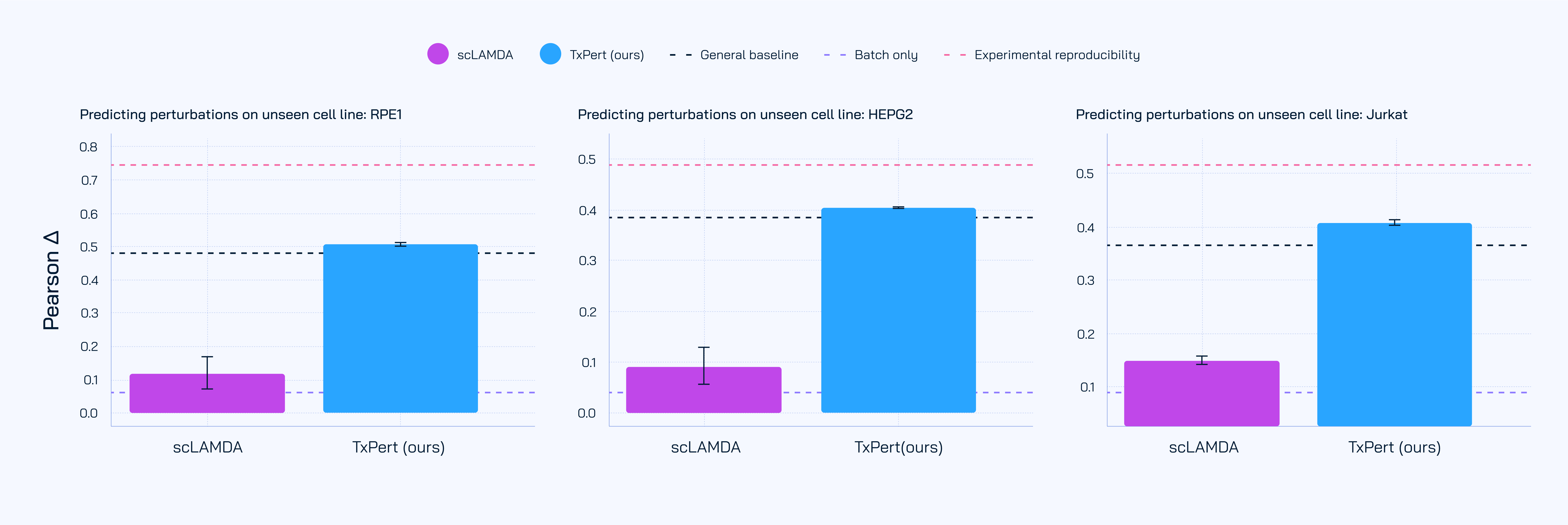}

  \caption{The performance of scLAMBDA and \modelname~at the task of 
    single perturbations in unseen cell lines. Horizontal bars put the predictions in the context of the general baseline,
    a learned model making predictions on the basis of batch information (in case of confounding between batch and perturbation), and an experimental
    reproducibility estimate.}
  \label{fig:xcell_sup}
\end{figure}

\begin{figure}[h]
  \centering
    \includegraphics[width=\textwidth]{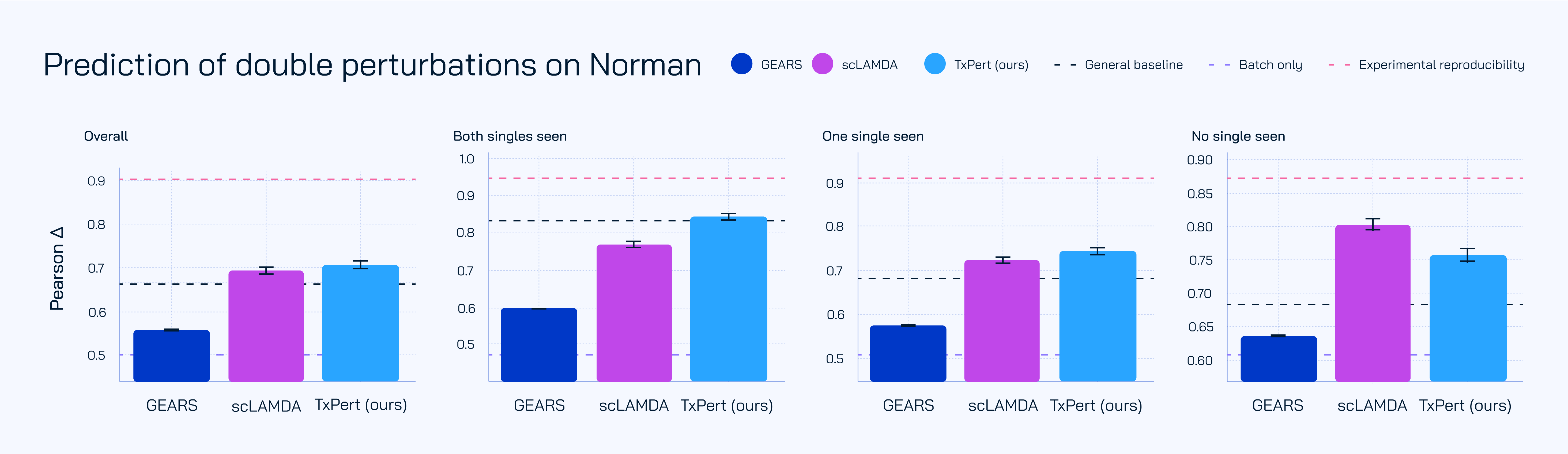}

  \caption{The performance of GEARS, scLAMBDA and \modelname~for prediction of double perturbations on Norman for various OOD settings. We report overall Pearson $\Delta$ and performance when both, one or none of the individual perturbations that constitute the double have been seen during training (left to right). Horizontal bars put the predictions in the context of the general baseline, a learned model making predictions on the basis of batch information (in case of confounding between batch and perturbation), and an experimental reproducibility estimate.}
  \label{fig:doubles-combined}
\end{figure}

\begin{figure}[h!]
    \centering
      \includegraphics[width=\textwidth]{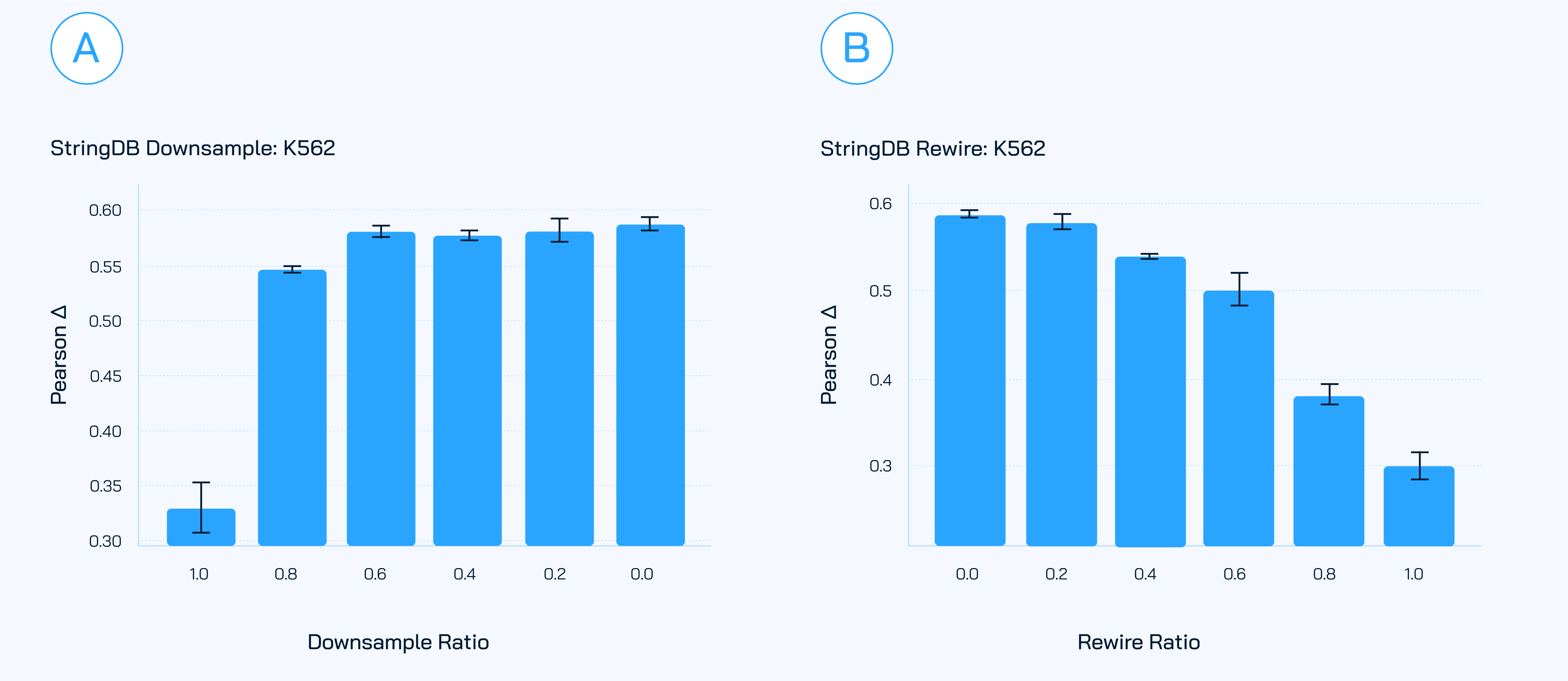}

    \caption{Perturbation effect prediction of unseen perturbations in K562: A) Randomly downsampling a fraction of the edges of the STRINGdb knowledge graph. In this experiment we randomly keep a ratio of the edges in the graph and remove the rest and analyze the performance of the method as we use more edges from the graph. B) Randomly rewiring a fraction of the source nodes in the STRINGdb graph. This experiment makes sure the edited graph still has similar in-degree of the original graph, but randomly rewire the sources each node receives information from.%
    }
    \label{fig:downsample_rewire}
\end{figure}

\begin{figure}[tb]

\centering

\includegraphics[width=\textwidth]{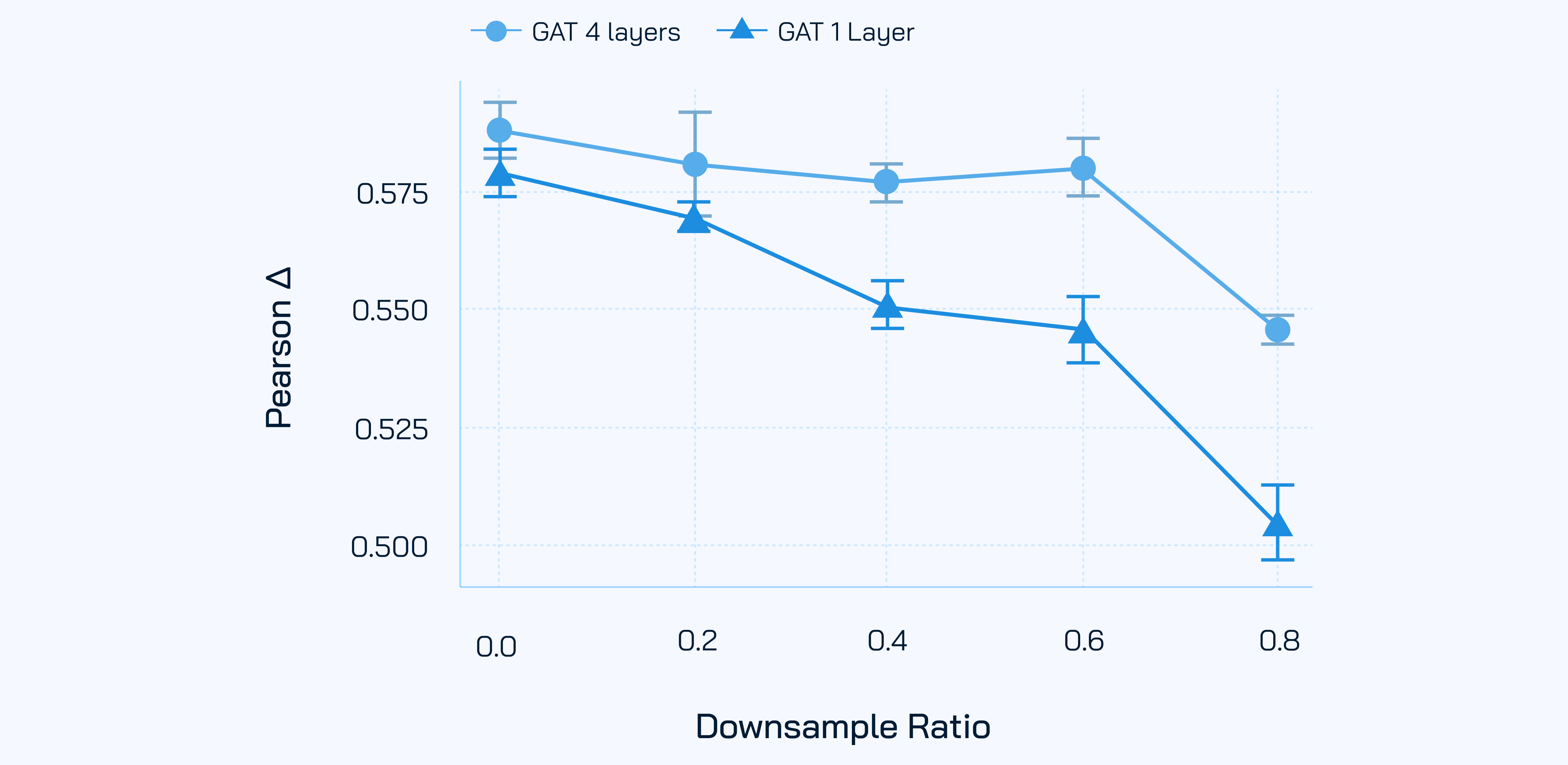}

\caption{Perturbation effect prediction of unseen perturbations in K562: This plot compares a 4-layer GAT model with a parameter-matched 1-layer GAT under varying edge downsample ratios. As shown in Figure~\ref{fig:downsample_rewire}A, the 4-layer model maintains stable performance across downsample ratios from 0.2 to 0.6, with a less pronounced drop than in the rewiring cases (Figures~\ref{fig:downsample_rewire}A and~\ref{fig:graph_ablation_main}A). We hypothesize that multi-hop reasoning allows the deeper model to compensate for removed edges—an advantage the 1-layer model lacks. To ensure a fair comparison, we match model capacity by doubling the hidden size of the 1-layer GAT. Results confirm that the 4-layer model is more robust to edge removal, showing a smaller performance decline.}

\label{fig:4vs1-layer-gnn}

\end{figure}

\begin{figure}[h!]
    \centering
    \includegraphics[width=0.7\textwidth]{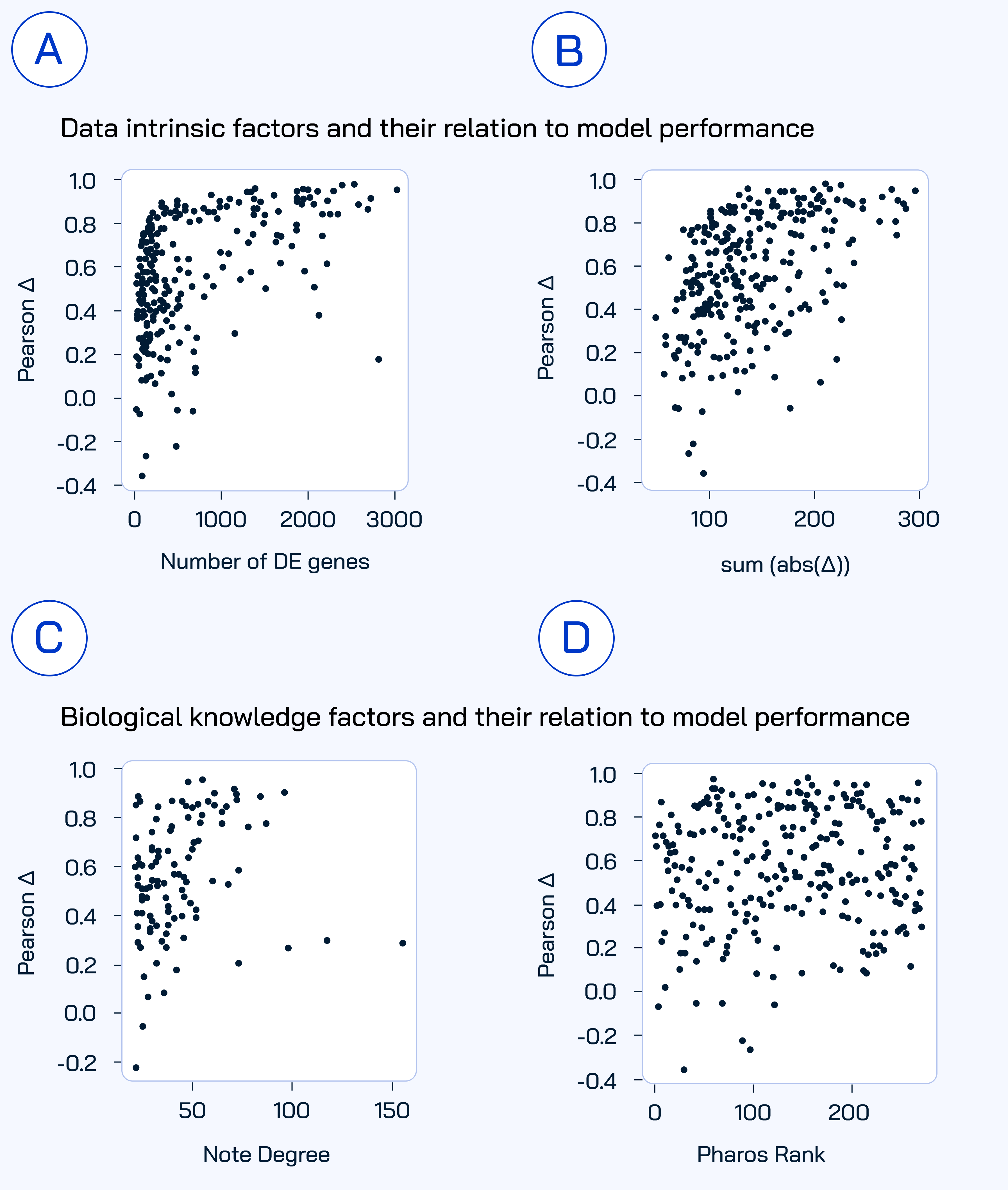}

    \caption{Exploration into potential relationships between perturbation-target metadata and the resulting performance
    as measured by Pearson $\Delta$, of \modelname. A) relation to the number of differentially expressed genes, a proxy for perturbation
    effect size, under the target perturbation
    (Wilcoxon, as implemented by scanpy). B) the relation to the sum of the absolute deltas, another proxy for the perturbation effect size,
    under the target perturbation. C) the relation to the degree of the perturbation target in the STRINGdb graph. D) the relation to the
    pharos knowledge rank (Section~\ref{sec:data}).}
    \label{fig:extra_eval}
\end{figure}

\renewcommand{\arraystretch}{1.5} %
{\footnotesize
\begin{longtable}{| c p{2.7cm} ccccccc |}
\caption{Significant functional GO enrichments (Holm corrected $p < 0.05$)in the (foreground; fg) gene set where the weighted average perturbation response (mean baseline) across all perturbations and datasets is $\bar{\Delta} > 0.05$. Background (bg) is the intersected of all reported assayed genes across datasets that's not in the foreground. Cat stands for ``category'', as in the genes with the given ``GO term'' assigned to them. The enrichment (e) ratio is $\frac{\text{cat fg} / \text{cat bg}}{\text{not cat fg} / \text{not cat bg}}$. Enrichments calculated with GOATOOLS.}\label{tab:go_mb_up}\\
    
    \hline
    GO term & description & cat fg & cat bg & not cat fg & not cat bg & e ratio & p & p holm \\
    \hline
    \endfirsthead
    
    \multicolumn{9}{c}%
    {{\tablename\ \thetable{} -- continued from previous page}} \\
    \hline
    GO term & description & cat fg & cat bg & not cat fg & not cat bg & e ratio & p & p holm \\
    \hline
    \endhead
    
    \hline \multicolumn{9}{|r|}{{Continued on next page}} \\ \hline
    \endfoot
    
    \hline
    \endlastfoot

    GO:0005773 & vacuole & 17 & 122 & 134 & 6079 & 6.32 & 2.29e-08 & 3.44e-04 \\
    GO:0000421 & autophagosome membrane & 7 & 12 & 144 & 6189 & 25.07 & 1.48e-07 & 2.23e-03 \\
    GO:0000323 & lytic vacuole & 14 & 95 & 137 & 6106 & 6.57 & 2.31e-07 & 3.48e-03 \\
    GO:0005764 & lysosome & 14 & 95 & 137 & 6106 & 6.57 & 2.31e-07 & 3.48e-03 \\
    GO:0019885 & antigen processing and presentation of endogenous peptide antigen via MHC class I & 4 & 1 & 147 & 6200 & 168.71 & 1.51e-06 & 2.27e-02 \\
    GO:0002483 & antigen processing and presentation of endogenous peptide antigen & 4 & 1 & 147 & 6200 & 168.71 & 1.51e-06 & 2.27e-02 \\
    
\end{longtable}

\begin{longtable}{| c p{2.7cm} ccccccc |}
    \caption{Top 20 significant functional GO enrichments foreground gene set where the weighted average perturbation response (mean baseline) across all perturbations and datasets is $\bar{\Delta} < -0.05$. See Tab\ref{tab:go_mb_up} for other details.}
    \label{tab:go_mb_down}\\
    
    \hline
    GO term & description & cat fg & cat bg & not cat fg & not cat bg & e ratio & p & p holm \\
    \hline
    \endfirsthead
    
    \multicolumn{9}{c}%
    {{\tablename\ \thetable{} -- continued from previous page}} \\
    \hline
    GO term & description & cat fg & cat bg & not cat fg & not cat bg & e ratio & p & p holm \\
    \hline
    \endhead
    
    \hline \multicolumn{9}{|r|}{{Continued on next page}} \\ \hline
    \endfoot
    
    \hline
    \endlastfoot
    
    GO:0003723 & RNA binding & 453 & 731 & 877 & 4291 & 3.03 & 2.59e-53 & 3.91e-49 \\
    GO:0034641 & cellular nitrogen compound metabolic process & 563 & 1240 & 767 & 3782 & 2.24 & 5.16e-35 & 7.77e-31 \\
    GO:0003676 & nucleic acid binding & 580 & 1341 & 750 & 3681 & 2.12 & 1.99e-31 & 3.00e-27 \\
    GO:0022402 & cell cycle process & 233 & 334 & 1097 & 4688 & 2.98 & 2.59e-30 & 3.90e-26 \\
    GO:0006725 & cellular aromatic compound metabolic process & 526 & 1181 & 804 & 3841 & 2.13 & 4.59e-30 & 6.92e-26 \\
    GO:1901360 & organic cyclic compound metabolic process & 543 & 1239 & 787 & 3783 & 2.11 & 8.20e-30 & 1.24e-25 \\
    GO:0046483 & heterocycle metabolic process & 525 & 1192 & 805 & 3830 & 2.1 & 6.53e-29 & 9.84e-25 \\
    GO:0006139 & nucleobase-containing compound metabolic process & 508 & 1139 & 822 & 3883 & 2.11 & 8.64e-29 & 1.30e-24 \\
    GO:1990904 & ribonucleoprotein complex & 225 & 331 & 1105 & 4691 & 2.89 & 3.28e-28 & 4.93e-24 \\
    GO:1901363 & heterocyclic compound binding & 728 & 1915 & 602 & 3107 & 1.96 & 2.13e-27 & 3.20e-23 \\
    GO:0097159 & organic cyclic compound binding & 731 & 1929 & 599 & 3093 & 1.96 & 3.35e-27 & 5.04e-23 \\
    GO:0051276 & chromosome organization & 125 & 124 & 1205 & 4898 & 4.1 & 1.38e-25 & 2.07e-21 \\
    GO:0044237 & cellular metabolic process & 761 & 2076 & 569 & 2946 & 1.9 & 6.31e-25 & 9.50e-21 \\
    GO:0090304 & nucleic acid metabolic process & 445 & 1000 & 885 & 4022 & 2.02 & 3.42e-24 & 5.15e-20 \\
    GO:0043228 & non-membrane-bounded organelle & 431 & 960 & 899 & 4062 & 2.03 & 7.80e-24 & 1.17e-19 \\
    GO:0043232 & intracellular non-membrane-bounded organelle & 431 & 960 & 899 & 4062 & 2.03 & 7.80e-24 & 1.17e-19 \\
    GO:0043933 & protein-containing complex organization & 340 & 696 & 990 & 4326 & 2.13 & 9.31e-23 & 1.40e-18 \\
    GO:0006259 & DNA metabolic process & 195 & 302 & 1135 & 4720 & 2.69 & 2.20e-22 & 3.31e-18 \\
    GO:0005654 & nucleoplasm & 630 & 1652 & 700 & 3370 & 1.84 & 4.98e-22 & 7.49e-18 \\
    GO:1903047 & mitotic cell cycle process & 175 & 256 & 1155 & 4766 & 2.82 & 6.71e-22 & 1.01e-17 \\
    GO:0032991 & protein-containing complex & 728 & 2031 & 602 & 2991 & 1.78 & 1.38e-20 & 2.08e-16 \\
\end{longtable}

\begin{longtable}{| c p{2.7cm} ccccccc |}
    \caption{Significant functional GO enrichments (Holm corrected $p < 0.05$)in the (foreground; fg) gene set comprised of the 50 perturbation targets in the test set 
    with the highest Pearson $\Delta$. Background (bg) is all non-foreground perturbation targets in the test set. 
    Cat stands for ``category'', as in the genes with the given ``GO term'' assigned to them.
    The enrichment (e) ratio is $\frac{\text{cat fg} / \text{cat bg}}{\text{not cat fg} / \text{not cat bg}}$. Enrichments calculated with GOATOOLS.}\label{tab:go_targ_best}\\
    
    \hline
    GO term & description & cat fg & cat bg & not cat fg & not cat bg & e ratio & p & p holm \\
    \hline
    \endfirsthead
    
    \multicolumn{9}{c}%
    {{\tablename\ \thetable{} -- continued from previous page}} \\
    \hline
    GO term & description & cat fg & cat bg & not cat fg & not cat bg & e ratio & p & p holm \\
    \hline
    \endhead
    
    \hline \multicolumn{9}{|r|}{{Continued on next page}} \\ \hline
    \endfoot
    
    \hline
    \endlastfoot
    
    GO:0090150 & establishment of protein localization to membrane & 13 & 4 & 37 & 218 & 19.15 & 8.91e-08 & 2.91e-04 \\
    GO:0006613 & cotranslational protein targeting to membrane & 12 & 4 & 38 & 218 & 17.21 & 4.61e-07 & 1.51e-03 \\
    GO:0006612 & protein targeting to membrane & 12 & 4 & 38 & 218 & 17.21 & 4.61e-07 & 1.51e-03 \\
    GO:0006614 & SRP-dependent cotranslational protein targeting to membrane & 12 & 4 & 38 & 218 & 17.21 & 4.61e-07 & 1.51e-03 \\
    GO:0072657 & protein localization to membrane & 13 & 6 & 37 & 216 & 12.65 & 7.61e-07 & 2.48e-03 \\
    GO:0006413 & translational initiation & 14 & 8 & 36 & 214 & 10.4 & 9.80e-07 & 3.20e-03 \\
    GO:0072594 & establishment of protein localization to organelle & 15 & 10 & 35 & 212 & 9.09 & 1.08e-06 & 3.52e-03 \\
    GO:0045047 & protein targeting to ER & 12 & 5 & 38 & 217 & 13.71 & 1.35e-06 & 4.42e-03 \\
    GO:0070972 & protein localization to endoplasmic reticulum & 12 & 5 & 38 & 217 & 13.71 & 1.35e-06 & 4.42e-03 \\
    GO:0019083 & viral transcription & 12 & 5 & 38 & 217 & 13.71 & 1.35e-06 & 4.42e-03 \\
    GO:0072599 & establishment of protein localization to endoplasmic reticulum & 12 & 5 & 38 & 217 & 13.71 & 1.35e-06 & 4.42e-03 \\
    GO:0051668 & localization within membrane & 13 & 7 & 37 & 215 & 10.79 & 1.88e-06 & 6.12e-03 \\
    GO:0033365 & protein localization to organelle & 15 & 12 & 35 & 210 & 7.5 & 4.33e-06 & 1.41e-02 \\
    GO:0006364 & rRNA processing & 19 & 22 & 31 & 200 & 5.57 & 5.62e-06 & 1.83e-02 \\
    GO:0000184 & nuclear-transcribed mRNA catabolic process, nonsense-mediated decay & 12 & 7 & 38 & 215 & 9.7 & 8.22e-06 & 2.67e-02 \\
    GO:0016072 & rRNA metabolic process & 19 & 23 & 31 & 199 & 5.3 & 8.98e-06 & 2.92e-02 \\
    GO:0006886 & intracellular protein transport & 16 & 16 & 34 & 206 & 6.06 & 1.21e-05 & 3.93e-02 \\
\end{longtable}

}

\end{document}